%% file: main.tex
\documentclass[sigconf, 10pt]{acmart}
\usepackage{bbding}
\usepackage{amsmath}
\usepackage{color}
\usepackage{dsfont}
\usepackage{multirow}
\usepackage{makecell}
\usepackage{caption}
\usepackage{subcaption}
\usepackage{enumitem}
\usepackage{algorithm}  
\usepackage{algpseudocode}  
\usepackage{amsmath}  
\usepackage[symbol]{footmisc}
\usepackage{pifont}
\usepackage{colortbl} 
\usepackage{bm}
\usepackage{xcolor}   
\usepackage{soul}
\usepackage{url}
\usepackage{booktabs}
\usepackage{tabularx}

\usepackage[most]{tcolorbox}

\definecolor{ao}{rgb}{0.0, 0.5, 0.0}
\definecolor{britishracinggreen}{rgb}{0.0, 0.26, 0.15}
\definecolor{carnelian}{rgb}{0.7, 0.11, 0.11}

\newcommand{\myparagraph}[1]{\vspace{1mm} \noindent \textbf{\textsf{#1}}}

\newcommand{\cmark}{{\color{ao} \ding{52}}}%
\newcommand{\xmark}{{\color{carnelian} \ding{56}}}%

\setcopyright{none}
\renewcommand\footnotetextcopyrightpermission[1]{} 
\begin{document}
\newcommand{\workname}{SensorPersona}

\title{\workname: An LLM-Empowered System for Continual Persona Extraction from Longitudinal Mobile Sensor Streams}

\author{
Bufang Yang$^{1}$, Lilin Xu$^{2}$, Yixuan Li$^{1}$, Kaiwei Liu$^{1}$, Xiaofan Jiang$^{2}$, Zhenyu Yan$^{1}$}

\affiliation{%
\institution{$^{1}$ The Chinese University of Hong Kong
\country{Hong Kong SAR},\\ 
$^{2}$ Columbia University
\country{United States}
}
}

\begin{abstract}
Personalization is essential for Large Language Model (LLM)-based agents to adapt to users’ preferences and improve response quality and task performance. However, most existing approaches infer personas from chat histories, which capture only self-disclosed information rather than users’ everyday behaviors in the physical world, limiting the ability to infer comprehensive user personas.
In this work, we introduce \workname, an LLM-empowered system that continuously infers stable user personas from multimodal longitudinal sensor streams unobtrusively collected from users’ mobile devices.
\workname~first performs person-oriented context encoding on continuous sensor streams to enrich the semantics of sensor contexts.
It then employs hierarchical persona reasoning that integrates intra- and inter-episode reasoning to infer personas spanning physical patterns, psychosocial traits, and life experiences.
Finally, it employs clustering-aware incremental verification and temporal evidence-aware updating to adapt to evolving personas.
We evaluate \workname~on a self-collected dataset containing 1,580 hours of sensor data from 20 participants, collected over up to 3 months across 17 cities on 3 continents.
Results show that \workname~achieves up to 31.4\% higher recall in persona extraction, an 85.7\% win rate in persona-aware agent responses, and notable improvements in user satisfaction compared to state-of-the-art baselines.

\end{abstract}

\settopmatter{printfolios=true}
\maketitle
\pagestyle{plain}

\input{secs/1_intro}
\input{secs/2_related_works}
\input{secs/3_motivation}
\input{secs/4_system}
\input{secs/5_evaluation}
\input{secs/6_discussion}

\vspace{-.5em}
\section{Conclusion}
We introduce \workname, an LLM-driven system that infers stable user personas from multimodal longitudinal sensor streams. 
\workname~employs persona-oriented context encoding and hierarchical reasoning to derive multidimensional personas, and incorporates persona maintenance to adapt to evolving personas.
Evaluations show that \workname~outperforms state-of-the-art baselines.




\bibliographystyle{ACM-Reference-Format}
\bibliography{main}

\clearpage

\appendix

\section*{Appendix}

\section{Details of User Study}

\begin{figure}[h]
    \centering
    \begin{subfigure}{0.49\columnwidth}  
    \centering \includegraphics[width=1.0\columnwidth]{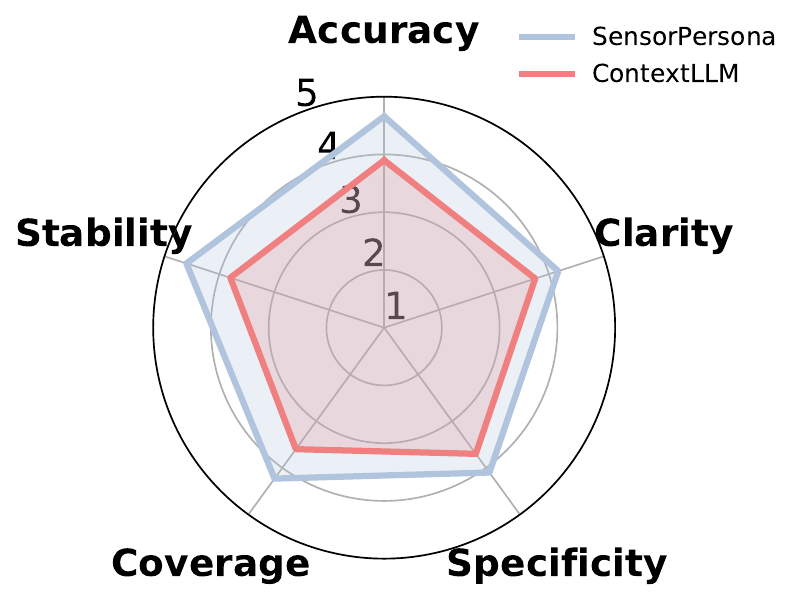}
            \vspace{-1.em} 
    \caption{All users} 
    \label{fig:user_study_1}
    \end{subfigure}
    \hfill
     \begin{subfigure}{0.49\columnwidth}
        \centering
        \includegraphics[width=1\columnwidth]{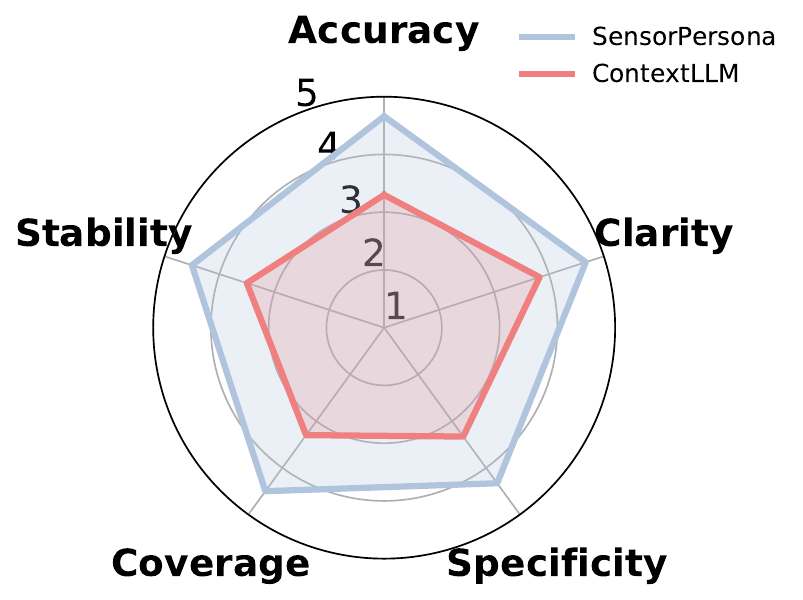}
        \vspace{-1.em} 
        \caption{Users with >100h data}
        \label{fig:user_study_100_1}
    \end{subfigure}
   \vspace{-0.8em} 
    \caption{Persona extraction results.}
\label{fig:user_study_100_E}
\vspace{-1.5em} 
\end{figure}

\begin{figure}[h]
    \centering
    \begin{subfigure}{0.49\columnwidth}
        \centering
        \includegraphics[width=1\columnwidth]{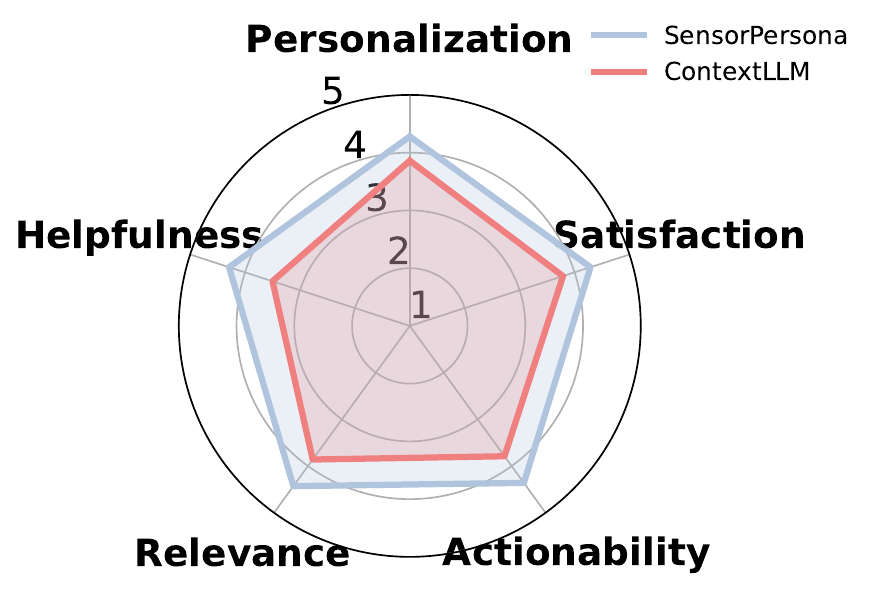}
        \vspace{-1.em} 
        \caption{All users}
        \label{fig:user_study_app_2}
    \end{subfigure}
    \hfill
    \begin{subfigure}{0.50\columnwidth}
        \centering
        \includegraphics[width=1\columnwidth]{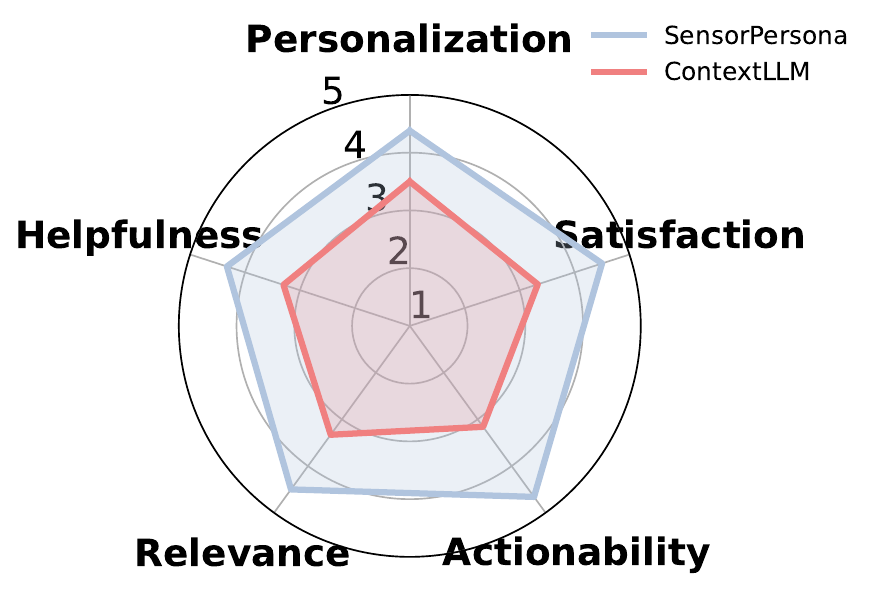}
        \vspace{-1.em} 
        \caption{Users with >100h data}
        \label{fig:user_study_100_2}
    \end{subfigure}
   \vspace{-2.em} 
    \caption{Agent response results.}
\label{fig:user_study_100_R}
\vspace{-1.em} 
\end{figure}

\begin{figure}[h]
    \centering
    \begin{subfigure}{\columnwidth}
        \centering
        \includegraphics[width=1\columnwidth]{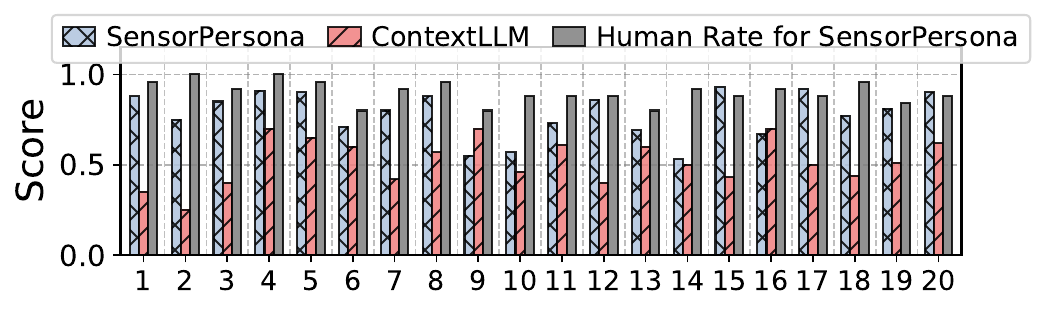}
        \vspace{-2em} 
        \caption{\textit{Recall} score comparison.}
        \label{fig:user_study_bar_recall}
    \end{subfigure}
    \hfill
    \begin{subfigure}{\columnwidth}
        \centering
        \includegraphics[width=1\columnwidth]{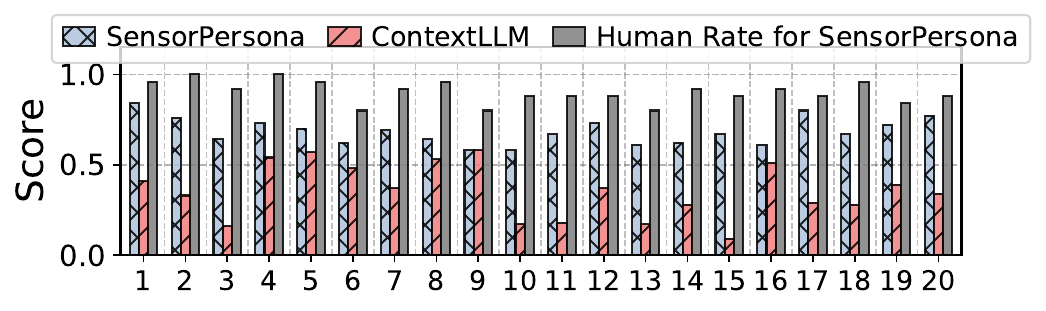}
        \vspace{-2em} 
        \caption{\textit{F1} score comparison.}
        \label{fig:user_study_bar_f1}
    \end{subfigure}
   \vspace{-2.em} 
    \caption{
    Comparison of the average \textit{Recall} and \textit{F1} across all participants for \workname, ContextLLM, and human reference rates for \workname.}
\label{fig:user_study_bar_comparison}
\end{figure}

Fig.~\ref{fig:user_study_100_E} and Fig.~\ref{fig:user_study_100_R} demonstrate the performance of \workname~based on human ratings from all users and from users with more than 100 hours of sensing data. Fig.~\ref{fig:user_study_100_E} shows that the average rating of \workname~increases from 4.35 across all users to 4.53 for users with more than 100 hours of data, while the baseline decreases from 3.70 to 3.45. Fig.~\ref{fig:user_study_100_R} shows a similar trend, where the average rating of \workname~increases from 4.33 to 4.48, while the baseline drops from 3.73 to 3.33. These results indicate that as sensing duration increases, the personas extracted by \workname~and the resulting agent responses achieve higher quality, whereas the baseline does not exhibit the same trend. This suggests that our approach more effectively leverages long-term behavioral signals when sufficient data is available.
Fig.~\ref{fig:user_study_bar_comparison} demonstrates \workname~consistently outperforms ContextLLM in both \textit{Recall} and \textit{F1} across users.

\begin{figure*}[t]
  \centering
\includegraphics[width=1\linewidth]{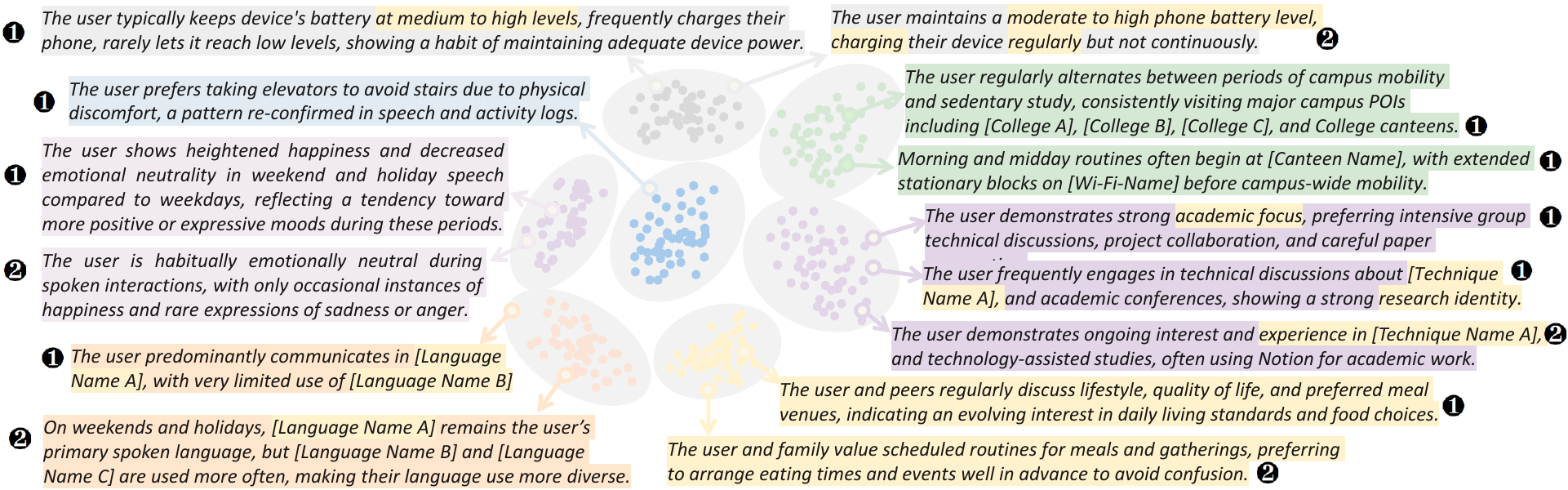}
\vspace{-2em}
  \caption{
  An example of persona clustering in \workname, where personas with similar patterns are grouped together despite variations across different time periods (Phase \ding{182} and Phase \ding{183}). Sensitive identifiers are anonymized.}
\label{fig:clustering_results}
\end{figure*}

\section{Dataset Details}
\label{appendix:dataset}
As shown in Tab.~\ref{tab:dataset_stats}, our evaluation dataset spans three months of multimodal sensing from 20 participants across 17 cities on three continents, totaling 1,580 hours of smartphone sensing data and providing rich longitudinal context across diverse users and environments.
We recruited 20 participants (10 male, 10 female) aged 18-63 years, ensuring diverse user coverage for robust persona extraction evaluation.
As for geographic diversity, data collection spanned 17 cities across three continents (Europe, North America, and Asia), including both major metropolitan areas and smaller cities.
During data collection, seven types of sensor data are captured, including audio recordings, IMU data, GPS, step count, network data (Wi-Fi SSID and cellular information), battery level, and screen brightness.
Participants used 13 different smartphones spanning multiple generations of iPhone series, Google Pixel 7, Huawei Mate40, and OnePlus Ace3 Pro, ensuring real-world applicability of our system.

The sensor cues used in this study are derived from multimodal sensor data and include the following types: battery level, battery state, screen brightness, location name, POIs (e.g., supermarket, shopping mall, convenience store, marketplace, commercial area, restaurant, bus station, subway station), user activity, network type, Wi-Fi SSID, language usage, emotion, speech content, and step count.

\section{Persona Clustering}
Fig.~\ref{fig:clustering_results} shows the clustering results of personas inferred from continual sensor streams by \workname. The participant experienced a significant change in location and lifestyle during the study. Specifically, in Phase 1 the participant studied and worked in one city, while in Phase 2 the participant spent about two weeks in another city during a vacation with family, resulting in noticeable changes in daily routines and preferences.
Results show that clustering still reveals many shared personas across phases, which are highlighted in the figure. Moreover, clustering naturally organizes different categories of personas, such as lifestyle preferences, daily routines, and occupational identities.

\vspace{-1.em}

\end{document}

%% file: secs/1_intro.tex
\section{Introduction}

\begin{figure}[t]
  \centering
\includegraphics[width=1\linewidth]{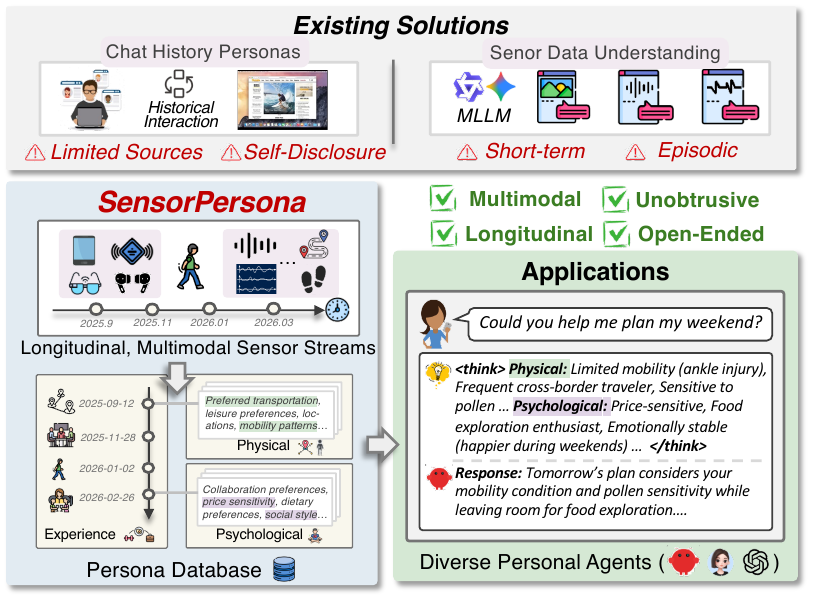}
\vspace{-2.em}
  \caption{Application scenario of \workname.}
\label{fig:teaser}
  \vspace{-1.em}
\end{figure}



Large Language Models (LLM)-based chatbots and agents have been increasingly popular in mobile agents~\cite{wang2024mobile}, web agents~\cite{wei2025webagent}, and personal assistants such as OpenClaw~\cite{OpenClaw}.
Personalization is crucial for aligning LLM agents with individual users’ preferences, such as tailoring response styles or task execution~\cite{cai2025large,xu2025mem}.
Such personalization often relies on \textit{personas}~\cite{ge2024scaling,gao2024aligning,wang2024ai}, which capture stable user traits such as preferences, interaction styles, and personal background.

Most existing studies derive personas for agent personalization primarily from chat histories and conversational interactions~\cite{xu2025mem,chhikara2025mem0,gao2024aligning}.
However, existing approaches primarily rely on explicit user \textit{self-disclosure} during conversations, while such textual interactions provide only a partial view of users’ personas.
In contrast, human personas are also reflected in users’ continuously evolving behaviors and interactions in the physical world rather than solely in conversational self-disclosures~\cite{APA_personality,haehner2024systematic}.
This highlights the importance of developing LLM-based systems that infer personas from longitudinal sensor streams.

Existing studies on sensor data understanding primarily focus on interpreting sensor signals using LLMs and Visual LLMs (VLMs)~\cite{han2024onellm,yang2024viassist}, while they primarily focus on short-term sensor interpretation.
Recent work also leverages LLMs to reason over multimodal sensor streams for activity logging~\cite{tian2025dailyllm}, journaling~\cite{xu2025autolife}, and sense-making~\cite{choube2025gloss,li2025vital}. 
However, these approaches focus on episodic event descriptions rather than inferring and maintaining longitudinal user personas.
Although earlier work has explored inferring human traits or behavioral patterns from sensor data~\cite{srinivasan2014mobileminer,wang2018sensing}, they primarily focus on prediction or pattern discovery, typically relying on task-specific feature engineering, closed-set models, or pattern mining techniques with limited generalization.
Deriving stable, long-term personas from continuous multimodal sensor streams therefore remains an open problem.

To bridge this research gap, we develop a system that infers user personas from longitudinal sensor streams, capturing persistent behavioral patterns during users’ interactions with the physical world. 
However, several unique challenges remain in developing such a system.
\textbf{First,} user personas reflect persistent behavioral tendencies and preferences that recur across weeks or months, rather than transient events~\cite{fleeson2001toward}. 
Unlike prior work that focuses on episodic event understanding from sensor signals~\cite{han2024onellm,post2025contextllm,yu2025sensorchat,xu2025autolife}, persona inference requires consistent behavioral evidence over extended periods.
Deriving longitudinal personas that capture users’ physical and psychological traits from continuous sensor streams remains challenging.
\textbf{Second,} inferring longitudinal personas requires LLM reasoning over continuous, multimodal sensor streams. 
Unlike semantically rich chat dialogues, always-on sensor data is highly redundant and semantically sparse, introducing substantial noise for reasoning. 
Existing long-context compression and persona extraction approaches~\cite{pan2024llmlingua,jiang2023llmlingua} primarily focus on textual or conversational traces with dense semantics. 
Inferring stable behavioral tendencies from such data remains challenging.
\textbf{Third,} user personas evolve over time, requiring continuous maintenance rather than one-time inference~\cite{roberts2006patterns}. 
Prior work primarily focuses on understanding episodic sensor events or generating activity journals~\cite{tian2025dailyllm,xu2025autolife}, rather than tracking personas over time. 
Continuously adapting to evolving personas from longitudinal sensor streams remains challenging.

In this study, we develop \workname, an LLM-driven system that infers stable user personas from longitudinal multimodal sensor streams unobtrusively collected by mobile devices, capturing persistent behavioral patterns in everyday interactions with the physical world, as illustrated in Fig.~\ref{fig:teaser}.
\workname~first employs a persona-oriented context encoding approach that converts massive streaming sensor data into semantically enriched sensor contexts.
Next, \workname~performs intra- and inter-episodic reasoning over these contexts to derive multidimensional personas that capture users’ physical and psychological patterns as well as life experiences. 
Finally, \workname~employs a hierarchical persona maintenance mechanism to adapt to evolving personas, using clustering-aware incremental verification for newly inferred personas and temporal evidence-aware updating to adjust persona weights.
The derived personas can serve as external memory, enabling diverse applications, including personal agents (e.g., OpenClaw~\cite{OpenClaw}) for personalized task execution~\cite{liu2025tasksense,liu2024tasking} and proactive agents~\cite{yang2025contextagent} for more accurate anticipation of user needs.

We evaluate \workname~on a self-collected real-world dataset from 20 participants’ daily lives, comprising 1,580 hours of smartphone sensor data that capture users’ natural interactions with the physical world.
We design three evaluation tasks to assess \workname, including persona extraction quality, persona-aware agent responses, and a user study with human ratings.
We implement \workname~on eight LLMs. Results show that \workname~achieves up to 31.4\% higher Recall and 25.9\% higher F1 in persona extraction. For the persona-aware agent response task, agents leveraging personas inferred by \workname~achieve up to an 85.7\% win rate over baselines in pairwise comparisons. 
In a user study, \workname~significantly outperforms baselines in human ratings, suggesting that the inferred personas align with participants’ self-perception while identifying personas users had not recognized.
We summarize the main contributions of this work as follows.

\begin{itemize}[leftmargin=*]


\item We introduce \workname, an LLM-empowered system that infers stable user personas from longitudinal multimodal sensor streams collected by mobile devices.

\item We develop a hierarchical persona reasoning approach that integrates intra- and inter-episodic reasoning to derive multidimensional personas capturing users’ physical and psychological traits and life experiences.

\item We design a persona-oriented context encoding approach and a hierarchical persona maintenance mechanism for streaming sensor data, enabling semantically enriched sensor contexts and dynamic adaptation to evolving personas.

\item We collect a real-world dataset with 1,580 hours of smartphone sensor data from 20 participants over up to 3 months, spanning 17 cities across 3 continents. Evaluation on three tasks, including a user study, shows that \workname~significantly outperforms state-of-the-art baselines.

\end{itemize}

\begin{table}
\footnotesize
\centering
\setlength{\tabcolsep}{5pt}
\caption{Comparison of \workname~with prior work.}
\vspace{-1.2em}
\label{tab:compare}
\resizebox{\columnwidth}{!}{
\begin{tabular}{c|ccccc}
\toprule
Methods & 
\begin{tabular}[c]{@{}c@{}}Multi-\\Modal\end{tabular}
 &
\begin{tabular}[c]{@{}c@{}}Long-\\Term\end{tabular}
 &
\begin{tabular}[c]{@{}c@{}}Open-\\Ended\end{tabular}
& 
\begin{tabular}[c]{@{}c@{}}System\\Outputs\end{tabular}
 &
\begin{tabular}[c]{@{}c@{}}System\\Setting\end{tabular} 

\\
\midrule
Mem0~\cite{chhikara2025mem0} &
\xmark &
\cmark &
\cmark & Personas & Chatbot \\


M3-Agent~\cite{long2025seeing} &
\cmark &
\xmark &
\cmark &Episodes &Camera \\

AutoLife~\cite{xu2025autolife}  &
\cmark &
\xmark  &
\cmark
& Episodes &
Smartphone\\

ContextLLM~\cite{post2025contextllm} &
\cmark &
\xmark  &
\cmark &
Episodes&
Wearables \\


MobileMiner~\cite{srinivasan2014mobileminer}  &
\cmark &
\cmark &
\xmark &
Behaviors&
Smartphone\\



\rowcolor[gray]{0.9}
\textbf{\workname} &
\cmark &
\cmark  &
\cmark &
Personas&
Smartphone\\
\bottomrule
\end{tabular}
}
\vspace{-1.5em}
\end{table}




%% file: secs/2_related_works.tex
\section{Related Works}

\myparagraph{Personalized LLM Agents}.
Prior works align LLMs with human preferences via one-off training on static, pre-collected annotations~\cite{ouyang2022training,rafailov2023direct}. 
Recent work shifts from one-time parametric personalization toward lifelong explicit personalization to reduce training overhead~\cite{zhong2024memorybank,chhikara2025mem0,liu2026simplemem}. 
They build memories or infer user personas from historical user–agent interactions and leverage them to adapt to users’ evolving preferences, such as writing style~\cite{gao2024aligning} and topical interests~\cite{ramos2024transparent,yang2025socialmind}.
Some studies explore lifelong personalization for LLMs while relying on synthetic personas rather than real-world profiles~\cite{wang2024ai,du2025twinvoice}. 
However, prior work relies on textual self-disclosure from chat interactions rather than unobtrusively capturing real-world behaviors through always-on, multimodal sensor streams, and thus cannot comprehensively characterize users’ physical and psychological personas.
Other work~\cite{yang2025contextagent,yang2025socialmind} leverages personas for personalized assistance but still depends on user-provided inputs rather than automatically inferring personas unobtrusively.

\myparagraph{Sensor Context Understanding}.
Recent studies have explored diverse approaches to leveraging LLMs and multimodal LLMs (MLLMs) to interpret sensor signals~\cite{post2025contextllm,yang2024drhouse,xu2024penetrative,yang2023edgefm}.
Other studies, such as AutoLife~\cite{xu2025autolife} and DailyLLM~\cite{tian2025dailyllm}, further generate daily journals or activity logs from sensor streams.
However, they primarily focus on understanding short-term sensor signals or logging episodic events, rather than inferring and maintaining longitudinal user personas.

\myparagraph{RAG-Based Sensemaking}.
Recent work provides interactive sensemaking for personal sensing data via question answering~\cite{choube2025gloss} and visualization~\cite{li2025vital}, typically using Retrieval-Augmented Generation (RAG) to retrieve relevant sensor contexts at query time.
Unlike episodic memories used in RAG systems, \workname~captures personas that represent stable user characteristics, providing an explicit user representation supporting diverse downstream agents.




\myparagraph{Longitudinal Human Behavior Modeling}.
Prior studies explore behavioral or health indicators from long-term mobile sensing, including daily routines~\cite{farrahi2008did}, mobility~\cite{zheng2008understanding}, stress~\cite{lu2012stresssense,wang2014studentlife}, academic performance~\cite{wang2015smartgpa}, and health prediction~\cite{xu2022globem,xu2023globem}.
However, they primarily rely on task-specific features and closed-set classifiers, limiting their ability to generalize to open-ended user personas.
Other work mines behavioral patterns from long-term mobile contexts or interaction traces~\cite{srinivasan2014mobileminer}. However, these approaches focus on predefined patterns or exemplar retrieval for prediction and execution rather than deriving open-ended user personas.
In contrast, \workname~infers stable, open-ended personas capturing both physical and psychosocial characteristics, serving as a user representation for personalized agents.

%% file: secs/3_motivation.tex
\vspace{-.5em}
\section{Background and Motivation}

\subsection{Background}
\subsubsection{User Personas}

Personality traits are typically described as relatively enduring patterns of thoughts, feelings, and behaviors that differentiate individuals~\cite{roberts2008personality}.
They reflect stable individual differences, exhibiting temporal stability and cross-situational consistency in everyday life~\cite{fleeson2001toward}, and can predict recurring tendencies and daily choices~\cite{ozer2006personality}.
Personas provide a concise representation of such stable user characteristics, reflecting recurring behavioral patterns, preferences, and interaction styles.
For example, ``\textit{consistently interested in dishes and snacks}'' reflects dietary preferences, 
``\textit{prefers elevators over stairs due to physical discomfort}'' indicates physical traits, 
and ``\textit{introverted and prefers staying at home on weekends}'' captures psychological characteristics.

\vspace{-.5em}
\subsubsection{Personal Agents}
Recent advances in LLMs have led to the emergence of increasingly capable personal assistant agents (e.g., OpenClaw~\cite{OpenClaw}), which can autonomously perform real-world tasks to assist users in everyday activities.
To provide personalized assistance, these agents must adapt to users’ preferences, routines, and interaction styles.

\myparagraph{Explicit Personalization}.
Many works align LLMs with human preferences via one-off training on pre-collected annotations~\cite{ouyang2022training,rafailov2023direct}. However, they rely on large-scale preference datasets, which are costly to obtain.
In contrast to static parametric fine-tuning, recent studies are shifting toward \emph{explicit personalization}~\cite{gao2024aligning,wang2024ai,chhikara2025mem0}, where agents infer personas from historical interactions and use them to adapt to evolving user preferences without repeatedly updating model weights.
They typically infer personas from historical dialogues and interaction trajectories within chatbot interfaces, yielding memories based on self-disclosed statements that reflect preferences, such as writing style and topical interests during chatbot interactions~\cite{gao2024aligning,ramos2024transparent,yang2025socialmind}.

\myparagraph{Applications Scenarios}.
During online execution, agents reference stored personas as personalized memory to generate responses and actions tailored to the user. Personas help LLMs interpret user intent and align responses with preferences, enabling personalized assistants for tasks such as content recommendation, writing, and social interaction~\cite{ramos2024transparent,gao2024aligning,yang2025socialmind}. They are also crucial for proactive agents~\cite{yang2025contextagent,yang2025proagent}, which anticipate user needs and provide timely assistance without explicit queries by leveraging environmental context and user personas. Personas thus serve as explicit user representations applicable across diverse personal agents.

\vspace{-.5em}
\subsection{Motivation}
\subsubsection{Limitations of Chatbot-Derived Personas}

\begin{figure}[t]
  \centering
\includegraphics[width=1\linewidth]{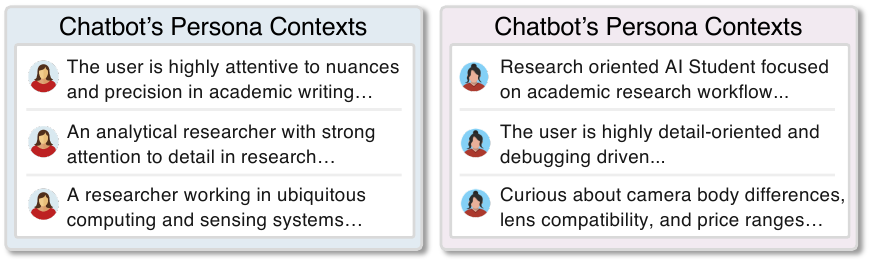}
\vspace{-2em}
 \caption{Example personas extracted from the chat histories of two users using ChatGPT memory.}
  \vspace{-1.2em}
\label{fig:motivation_chatbot}
\end{figure}

Recent studies have explored inferring user personas from conversational histories in chatbot interactions~\cite{gao2024aligning,ramos2024transparent,zhong2024memorybank}. As illustrated in Fig.~\ref{fig:motivation_chatbot}, personas stored in an individual’s chatbot memory (e.g., ChatGPT) typically capture \textit{communication preferences} (e.g., preferred interaction style), \textit{topical interests} (e.g., frequently discussed subjects), and \textit{self-disclosed cues} (e.g., personal information explicitly provided by the user). 
However, most existing studies on persona extraction for chatbots and LLM agents remain confined to what users explicitly express in conversation~\cite{xu2025mem,chhikara2025mem0}, inherently limited by self-disclosure and missing behavioral patterns evidenced in real-world interactions and multimodal sensor contexts.

In fact, personas should capture stable individual characteristics across time and situations, including physical and psychological aspects as well as life experiences~\cite{APA_personality,haehner2024systematic}.
However, many of these characteristics arise through recurring interaction in the physical world and can only be observed through longitudinal sensor data rather than within the chat interface alone~\cite{haehner2024systematic,wood2007new}.
This motivates us to develop a system that infers user personas from longitudinal sensor streams, capturing users’ recurring behavioral patterns during natural interactions with the physical world.


\begin{figure}[t]
  \centering
\includegraphics[width=1\linewidth]{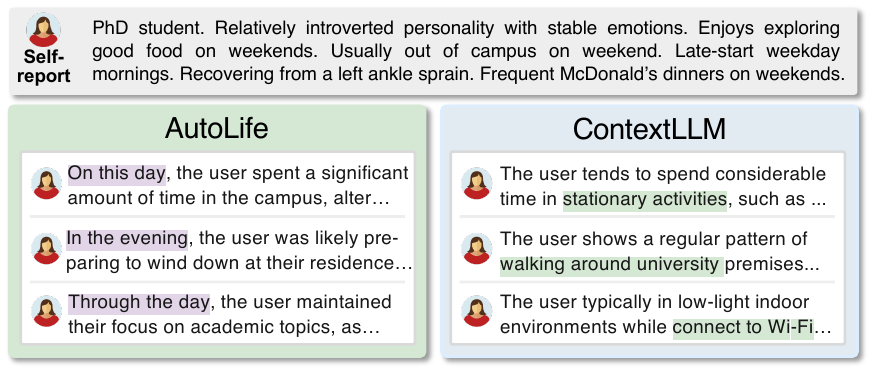}
\vspace{-2em}
  \caption{Performance of existing sensor context understanding approaches for persona inference.}
\label{fig:motivation_autolife}
  \vspace{-1.6em}
\end{figure}

\subsubsection{Inferring Personas from Longitudinal Sensor Streams}
\begin{figure}[t]
    \centering
    \begin{subfigure}{0.49\columnwidth}  
    \centering \includegraphics[width=1.0\columnwidth]{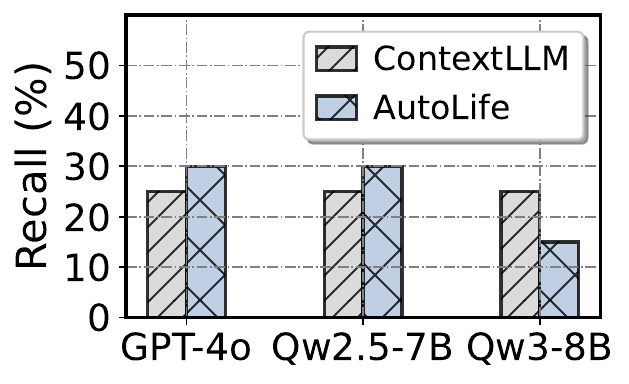}
    \end{subfigure}
    \hfill
     \begin{subfigure}{0.49\columnwidth}
        \centering
        \includegraphics[width=1\columnwidth]{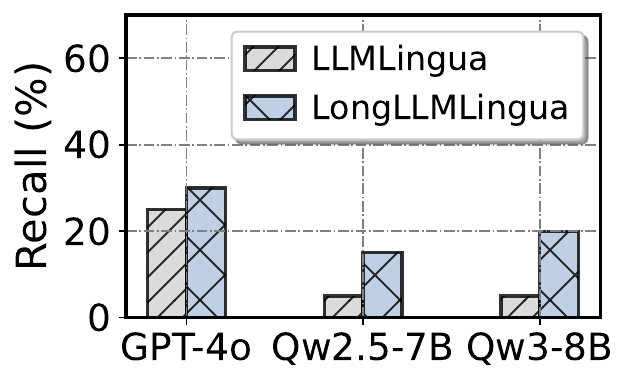}
    \end{subfigure}
   \vspace{-1em} 
    \caption{Performance of existing approaches for persona inference from longitudinal sensor streams.}
\label{fig:motivation_recall}
\vspace{-1em} 
\end{figure}

We first evaluate whether existing sensor data understanding approaches can infer stable personas from longitudinal sensor streams by examining two existing systems, AutoLife~\cite{xu2025autolife} and ContextLLM~\cite{post2025contextllm}. 
The evaluation is conducted on our self-collected dataset containing multimodal sensor streams (see \S~\ref{sec:data_collection}). 
The left panel in Fig.~\ref{fig:motivation_autolife} shows that AutoLife primarily generates daily narratives rather than longitudinal behavioral patterns across days, such as recurring routines and persistent preferences.
Although we prompt ContextLLM to focus on stable personas across days, long-term multimodal sensor streams spanning weeks introduce substantial noise and redundancy that hinder reasoning. 
Consequently, the inferred personas tend to capture coarse-grained statistics rather than recurring behavioral patterns and persistent preferences.
We use \textit{Recall} to measure how many self-reported personas are recovered. Fig.~\ref{fig:motivation_recall} shows that these approaches achieve only around 25.0\% \textit{Recall} in persona extraction.

We further evaluate whether existing long-context compression techniques can alleviate this issue by applying LLMLingua~\cite{pan2024llmlingua} and LongLLMLingua~\cite{jiang2024longllmlingua} to longitudinal sensor contexts. 
Fig.~\ref{fig:motivation_recall} illustrates that they still achieve limited \textit{Recall} in persona extraction. 
Their prompt compression strategies are designed for semantically rich natural language, where key information appears in explicit words and sentences. In contrast, longitudinal sensor contexts contain sparse and implicit cues across modalities and time, thus limiting prompt compression effectiveness for persona inference.

\subsubsection{Evolving Personas Over Time}

\begin{figure}[t]
    \centering
    \begin{subfigure}{0.48\columnwidth}  
    \centering \includegraphics[width=1\columnwidth]{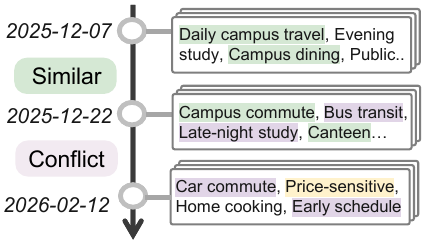}
            \vspace{-1.5em} 
    \caption{Examples of evolving personas over time.} \label{fig:motivation_streaming_example}
    \end{subfigure}
    \hfill
     \begin{subfigure}{0.48\columnwidth}
        \centering
        \includegraphics[width=1\columnwidth]{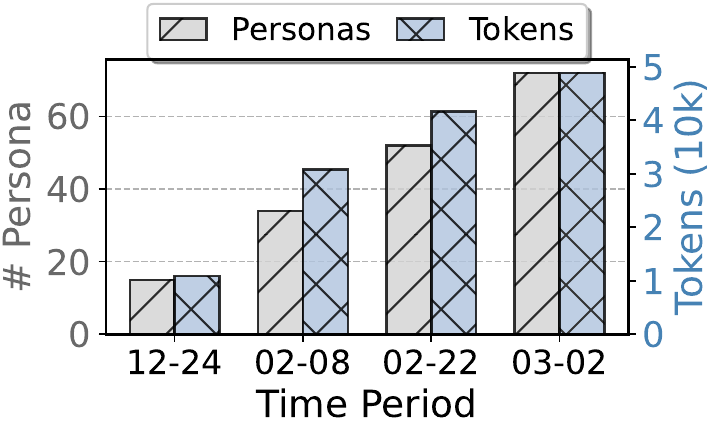}
        \vspace{-1.5em} 
        \caption{Persona expansion and token consumption over time.} \label{fig:motivation_streaming}
    \end{subfigure}
   \vspace{-1.em} 
    \caption{Evolving personas over time.}
\label{fig:motivation_streaming_}
\vspace{-1em} 
\end{figure}

Unlike short-term sensory data understanding, personas represent relatively persistent user characteristics that evolve over time~\cite{fleeson2001toward,roberts2006patterns}. 
When personas are inferred from longitudinal streaming sensor data, new personas may continuously emerge as more data arrives.
As users’ habits, environments, and behavioral patterns change over time, personas inferred from different periods may exhibit similarities or even conflict, as illustrated in Fig.~\ref{fig:motivation_streaming_example}.
However, continuously extracting personas from streaming sensor data causes their number to grow over time. Fig.~\ref{fig:motivation_streaming} shows the number of personas extracted from a user over three months in our self-collected dataset and the corresponding token consumption.
Results demonstrate that the number of personas can continually grow, increasing the token cost of conflict and redundancy checking and leading to substantial system overhead.

\vspace{-.5em}
\subsubsection{Summary}
Existing studies either derive personas from chat histories limited to self-disclosure or focus on interpreting short-term sensor signals to generate episodic descriptions. 
In contrast, \workname~infers stable personas that evolve over time from multimodal longitudinal sensor streams capturing persistent behavioral patterns in users’ interactions with the physical world.


%% file: secs/4_system.tex
\vspace{-.5em}
\section{System Design}

\begin{figure*}[t]
  \centering
\includegraphics[width=0.9\linewidth]{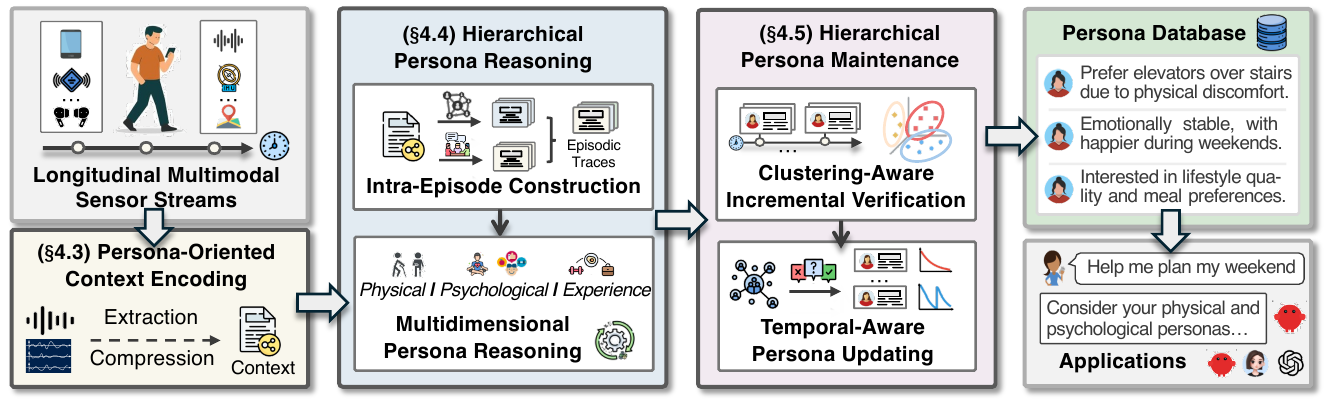}
\vspace{-1.em}
  \caption{System overview of \workname.}
\label{fig:overview}
  \vspace{-1.3em}
\end{figure*}

\subsection{Problem Formulation.}

We study the problem of maintaining stable user personas from longitudinal multimodal sensor streams.
Let $\mathcal{X}=\{x_t\}_{t=1}^{T}$ denote sensor streams continuously collected by mobile devices. 
The objective is to derive and maintain a set of personas $\mathbf{P}=\{\mathbf{P}_{\text{phy}},\mathbf{P}_{\text{psy}},\mathbf{E}\}$ from $\mathcal{X}$. 
Here $\mathbf{P}_{\text{phy}}$ represents recurring physical patterns (e.g., mobility routines), $\mathbf{P}_{\text{psy}}$ denotes psychosocial characteristics inferred from user interactions, and $\mathbf{E}$ stores the supporting evidence for each persona derived from episodic events. 
Specifically, each persona $p\in\mathbf{P}$ is represented as $p=(d_p,\mathcal{E}_p)$, where $d_p$ denotes the persona description and $\mathcal{E}_p$ records the episodes supporting persona $p$. 
As new sensor data arrive over time, the system incrementally updates $\mathbf{P}$ to capture persistent behavioral patterns while updating personas as they evolve.

\subsection{System Overview}
\workname~is an LLM-powered system that derives multidimensional personas reflecting users’ physical and psychological characteristics from multimodal, longitudinal sensor streams.
Fig.~\ref{fig:overview} demonstrates the system overview. \workname~utilizes sensor streams collected from mobile devices, capturing continuous behavioral patterns and interactions with the environment. It first applies persona-oriented context encoding (\S~\ref{sec:Persona-Oriented Context Encoding}) to transform raw sensor streams into multimodal sensor contexts with richer semantics. 
Next, \workname~performs hierarchical persona reasoning over sensor contexts, integrating intra- and inter-episodic reasoning to derive multidimensional personas (\S~\ref{sec:Hierarchical Multidimensional Persona Reasoning}).
Finally, it continuously infers personas from streaming sensor data and performs hierarchical persona maintenance (\S~\ref{sec:Hierarchical Persona Maintenance}), using clustering-aware incremental verification and temporal evidence-aware weighting to update personas over time.
The derived personas can support various downstream agent applications, such as LLM chatbots and personal assistants.

\vspace{-.5em}
\subsection{Persona-Oriented Context Encoding}
\label{sec:Persona-Oriented Context Encoding}

Personas reflect persistent user characteristics derived from recurring behavioral patterns rather than transient episodic events~\cite{fleeson2001toward}. Inferring such personas from multimodal sensor streams requires reasoning over longitudinal observations, posing additional challenges. To address this, we develop a persona-oriented context encoding approach that derives semantically enriched contexts from sensor streams.



\subsubsection{Multimodal Context Extraction}





\vspace{-.5em}
\workname~continuously collects low-cost sensor streams from mobile devices. 
Specifically, it records data from smartphones, such as audio, IMU, GPS, network status, and pedometer. 
Since raw sensor streams are difficult to interpret and not naturally aligned with the intrinsic knowledge of LLMs, \workname~leverages external models and tools to translate each modality into textual \textit{sensor cues}, as described below.



\textit{Audio.} \workname~derives speech-related cues, including spoken content, emotional tone, and language usage. 
It further determines whether each utterance is produced by the user or by others using the user’s voice fingerprint.


\textit{IMU.}
It employs a lightweight, customized neural network to classify the user’s motion activities from raw IMU signals.

\textit{Location.}
\workname~leverages Google Maps for reverse geocoding and to retrieve nearby points of interest (POIs) based on the user’s GPS coordinates. The search radius for POIs is set to 100 meters and includes categories such as supermarkets and transportation hubs.

\textit{Other Modalities.}
Sensor readings such as Wi-Fi SSID, cellular information, battery level, step counts, and screen brightness are used directly as cues without additional processing.


After obtaining sensor cues from each modality, \workname~synchronizes them to generate sensor contexts 
$\mathcal{S}=\{(\tau_t,\mathbf{s}_t)\}_{t=1}^{T}$, 
where $\tau_t$ is the timestamp at time step $t$, 
$\mathbf{s}_t=[s_{t,1},\ldots,s_{t,M}]$, 
and $s_{t,m}$ denotes the $m$-th sensor cue. 
$M$ denotes the total number of cues used in the system.

\begin{figure}
    \centering
    \begin{subfigure}{0.49\columnwidth}  
    \centering \includegraphics[width=1.0\columnwidth]{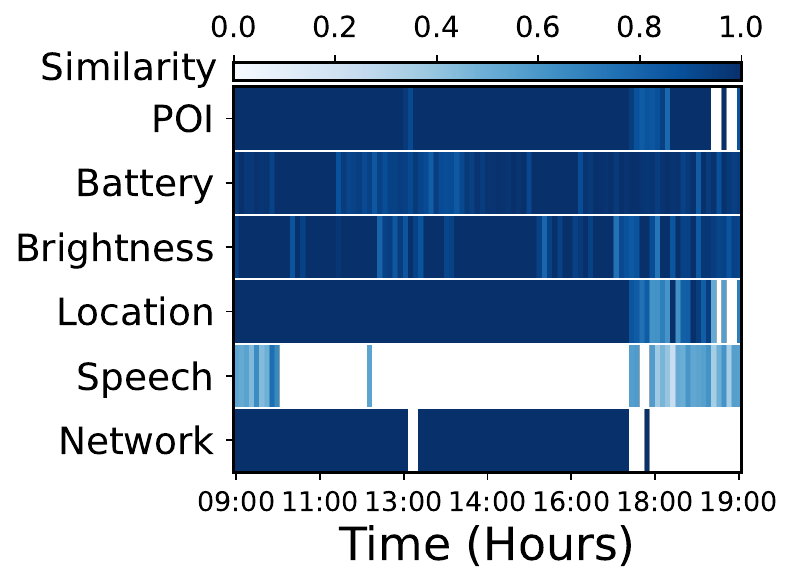}
            \vspace{-1.em} 
    \end{subfigure}
    \hfill
     \begin{subfigure}{0.49\columnwidth}
        \centering
        \includegraphics[width=1\columnwidth]{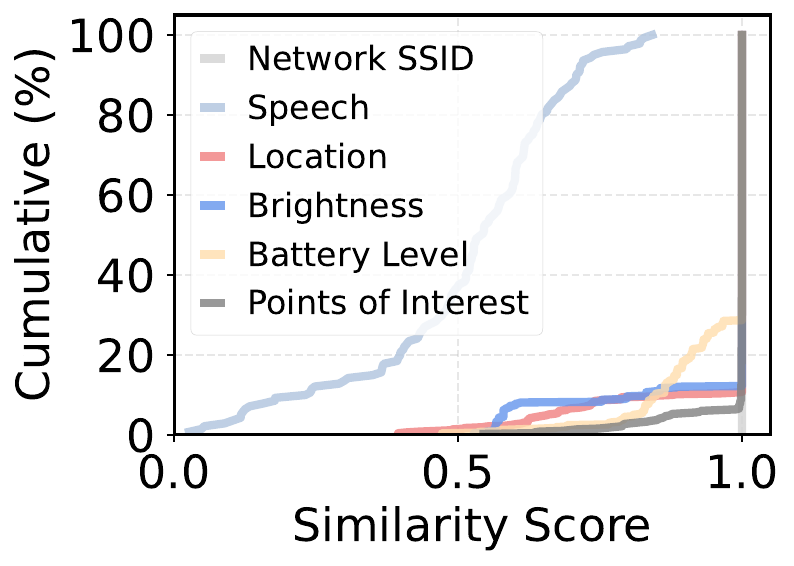}
        \vspace{-1.em} 
    \end{subfigure}
   \vspace{-1em} 
    \caption{Semantic similarity heatmap of multimodal sensor contexts over time (left) and the corresponding similarity score distribution (right), suggesting that similar patterns frequently occur.}
\label{fig:similarity_cdf_combined_2025-12-02}
\vspace{-1.5em} 
\end{figure}


\vspace{-.5em}
\subsubsection{Incremental Semantic Context Compression}
\label{sec:Incremental Semantic Context Compression}
The derived sensor contexts often exhibit substantial redundancy, with many cues reflecting minor fluctuations rather than substantive context changes, as shown in Fig.~\ref{fig:similarity_cdf_combined_2025-12-02}.
Reasoning over such redundant contexts with an LLM not only incurs substantial overhead but also hinders the discovery of persistent behavioral patterns. 
Although some recent work~\cite{ouyang2024llmsense} considers sensor context compression, it relies on heuristics for context selection, limiting generalizability. Meanwhile, existing long-context compression approaches focus on high-density textual dialogues rather than sensor streams~\cite{pan2024llmlingua,jiang2024longllmlingua}.

To address this, \workname~employs an incremental semantic context compression approach that leverages multidimensional sensor cues to identify informative contexts and transform $\mathcal{S}$ into a compact sequence of semantically coherent segments.
Specifically, to leverage complementary contextual information across sensor cues, \workname~derives context similarity by integrating multiple sensor cues. 
\workname~selects a cue subset $\mathcal{M}'\subset\mathcal{M}$ and forms a textual representation $x_t$ by concatenating the corresponding cue values,
$x_t=\texttt{concat}(\{s_{t,m}\mid m\in\mathcal{M}'\})$.
We then compute a semantic embedding $\mathbf{e}_t=f_{\text{emb}}(x_t)$ and incrementally merge contexts using cosine similarity
$\mathrm{sim}_t=\frac{\mathbf{e}_t^\top\mathbf{e}_{t-1}}{\|\mathbf{e}_t\|\,\|\mathbf{e}_{t-1}\|}$.
If $\mathrm{sim}_t\ge\alpha$, the current context is merged with the previous segment and the segment end time is updated to $\tau_t$.
For numeric sensor cues (e.g., brightness and battery level), values are averaged within each segment. 
For categorical cues (e.g., POIs and SSID), labels and their relative proportions are retained. 
Speech content is preserved across segments. 
\subsection{Hierarchical Multidimensional Persona Reasoning}
\label{sec:Hierarchical Multidimensional Persona Reasoning}
Recent studies primarily leverage LLMs to interpret short spans of sensor data or summarize episodic events~\cite{yang2026efficient,post2025contextllm,xu2025autolife,tian2025dailyllm,yu2025sensorchat}. 
In contrast, persona inference requires reasoning over multimodal sensor contexts across longitudinal timescales of weeks or months~\cite{fleeson2001toward} to identify recurring behavioral patterns that persist over time.
Although sensory contexts are substantially compressed in \S~\ref{sec:Incremental Semantic Context Compression}, their multimodal nature and long temporal span still introduce noise and pose challenges for LLM reasoning.
To address this challenge, we develop a hierarchical multidimensional persona reasoning approach that leverages both intra-episode and inter-episode reasoning to derive stable and comprehensive personas.

\vspace{-.5em}
\subsubsection{Intra-Episode Construction}


Following prior work in personality psychology, we characterize personas as recurring experiences and behavioral patterns~\cite{ozer2006personality}.
To capture such evidence, \workname~first derives episodic traces from sensor contexts, representing cues of individual experiences that support subsequent persona reasoning.

\begin{figure}[t]
  \centering
\includegraphics[width=1\linewidth]{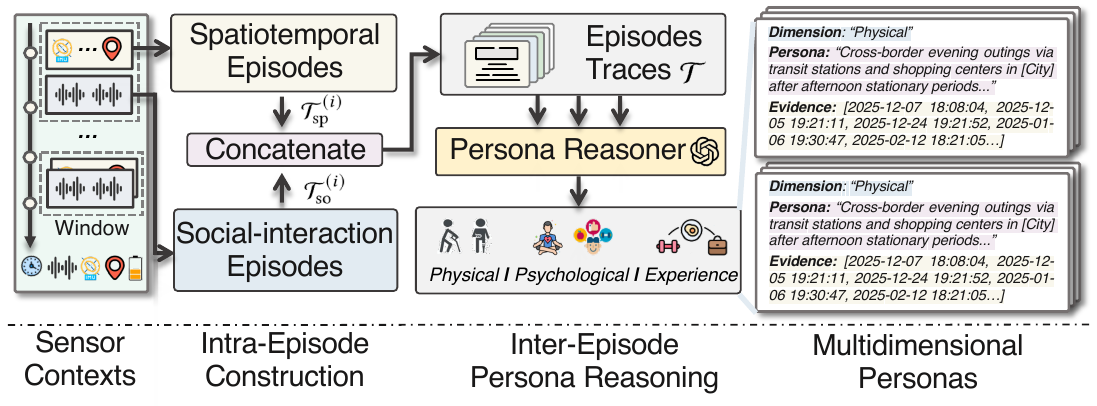}
\vspace{-2.1em}
\caption{\workname~derives personas from sensor contexts via inter- and intra-episode reasoning.}
\label{fig:persona_reasoning}
\end{figure}

Specifically, \workname~first segments sensor contexts into consecutive time windows of length $T$ and constructs episodic traces within each window, to reduce noise caused by the semantic sparsity of sensor data.
As shown in Fig.~\ref{fig:persona_reasoning}, for each window $i$, \workname~infers the episodic traces as
$(\mathcal{T}_{\text{sp}}^{(i)},\,\mathcal{T}_{\text{so}}^{(i)})
=\texttt{LLM}_{\texttt{episodic}}(p_{\text{episodic}},\,\mathcal{S}_{T}^{(i)}, \mathcal{K})$,
where $T$ denotes window length.
$\mathcal{T}_{\text{sp}}^{(i)}$ contains spatiotemporal episodes reflecting physical patterns (e.g., routines and mobility cues), while $\mathcal{T}_{\text{so}}^{(i)}$ contains social-interaction episodes inferred from audio.
$T$ is a hyperparameter that trades off the granularity of persona-relevant cues and system overhead (see \S~\ref{sec:Impact of Hyper-parameters}).
$\mathcal{K}$ is external knowledge used to improve context interpretation, such as calendar information for identifying weekends and holidays and web knowledge for interpreting contextual cues (e.g., Wi-Fi SSID semantics). 
To ensure traceability for subsequent persona reasoning, we use prompt $p_{\text{episodic}}$ to instruct the LLM to attach timestamps to each inferred episode. 
We then aggregate episodes from both dimensions across all windows:
$\mathcal{T}=\texttt{Concat}\big(\{\mathcal{T}_{\text{sp}}^{(i)}\}_{i=1}^{N},\ \{\mathcal{T}_{\text{so}}^{(i)}\}_{i=1}^{N}\big)$, each episode in $\mathcal{T}$ contains a textual description and its timestamp.

\vspace{-.5em}
\subsubsection{Inter-Episode Multidimensional Persona Reasoning}
\label{sec:Inter-Episode Multidimensional Persona Reasoning}

\workname~then leverages a persona reasoner to derive multidimensional personas from the extracted episodic traces:
$\mathbf{P}=\texttt{LLM}_{\texttt{reasoner}}(p_{\text{persona}}, \mathcal{T}, \mathcal{K})$.
Motivated by findings in personality psychology~\cite{roberts2008personality,fleeson2001toward,ozer2006personality}, \workname~captures personas along multiple dimensions, including physical patterns, psychosocial characteristics, and lifelong experiences, formalized as $\mathbf{P}=\{\mathbf{P}_{\text{phy}},\mathbf{P}_{\text{psy}},\mathbf{E}\}$.

Specifically, we prompt the LLM to derive physical and psychosocial personas, denoted as $\mathbf{P}_{\text{phy}}$ and $\mathbf{P}_{\text{psy}}$, each supported by a set of supporting episodes recorded in $\mathbf{E}$.
The physical persona $\mathbf{P}_{\text{phy}}$ captures recurring behavioral patterns and spatiotemporal regularities (e.g., frequent cross-border traveler).
The psychosocial characteristics $\mathbf{P}_{\text{psy}}$ capture the user’s social interaction style and psychosocial traits (e.g., price-sensitive).
To ensure traceability, each persona is grounded in multiple supporting episodes.
Each persona $p \in \mathbf{P}$ is represented as $p=(d_p,\mathcal{E}_p)$, 
where $d_p$ denotes the persona description and $\mathcal{E}_p$ denotes the traceable evidence, consisting of supporting episodes with their timestamps across repeated occurrences.

Note that personas reflect recurring and stable user characteristics~\cite{fleeson2001toward}.
Accordingly, physical patterns are extracted as personas only when they appear consistently across repeated observations, whereas psychological personas, such as preferences or identity-related information, can be promoted to personas more directly.
Finally, the extracted personas are treated as candidate personas, which are further processed and validated in \S~\ref{sec:Hierarchical Persona Maintenance} to determine subsequent maintenance.

\subsection{Hierarchical Persona Maintenance}
\label{sec:Hierarchical Persona Maintenance}
Unlike prior work that mainly focuses on short-horizon sensor signal understanding~\cite{tian2025dailyllm,post2025contextllm,han2024onellm}, personas evolve as life periods change and experiences accumulate, where behavioral and interaction patterns can shift over time~\cite{fleeson2001toward,roberts2006patterns}. 
Simply streaming persona inference can gradually lead to an explosion in the number of personas, while fine-grained similarity checking incurs significant system overhead.
To address this challenge, \workname~employs a hierarchical persona maintenance mechanism that dynamically adapts to such shifts while efficiently capturing evolving user patterns.


\begin{figure}[t]
  \centering
\includegraphics[width=0.92\linewidth]{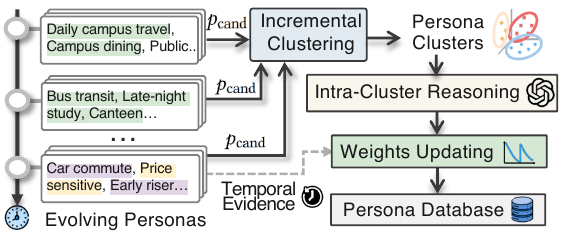}
\caption{Hierarchical persona maintenance in \workname~for adapting to evolving personas.}
\label{fig:persona_maintenance_v1}
\end{figure}

\subsubsection{Clustering-Aware Incremental Verification}



\workname~continuously derives candidate personas from incoming sensor streams. 
These personas capture multiple aspects of a user, such as lifestyle habits, social roles, mobility patterns, and social characteristics.
As persona extraction proceeds, the persona set grows in both scale and diversity, forming recurring persona patterns across different periods and contexts and posing challenges for efficient persona maintenance. 
To address this, \workname~first employs clustering-aware incremental verification to automatically discover shared persona patterns across varying periods and contexts, enabling efficient persona maintenance over time.



\myparagraph{Clustering-Aware Incremental Matching.}
Specifically, \workname~maintains a persona database $\mathcal{D}$ consisting of a set of persona clusters constructed incrementally over time. 
Each cluster is represented by a centroid embedding capturing the shared semantic pattern of personas within the cluster.
As shown in Fig.~\ref{fig:persona_maintenance_v1}, for each incoming candidate persona $p_{\text{cand}}$, \workname~performs incremental
clustering, computing its semantic similarity with the centroid of each existing cluster, and identifying the most similar cluster 
$c^{*}=\arg\max_{c\in\mathcal{D}}\mathrm{sim}(p_{\text{cand}},c)$. 
If the similarity exceeds a threshold $\theta$, the candidate persona is assigned to cluster $c^{*}$ and the centroid is updated accordingly. Otherwise, a new cluster is initialized with $p_{\text{cand}}$. 
We set $\theta$ to 0.65 based on the observed clustering performance, as shown in Fig.~\ref{fig:clustering_results}, where similar personas are grouped together even when behavioral patterns vary across different time periods.

\noindent\textbf{Intra-Cluster Semantic Reasoning.}
After clustering, \workname~performs intra-cluster semantic reasoning to maintain a concise and consistent set of personas. While embedding similarity is effective for coarse clustering, it cannot reliably determine whether two personas describe equivalent behavioral patterns or conflicting tendencies. To address this limitation, \workname~leverages an LLM-based semantic judge within each cluster.
For each candidate persona $p_{\text{cand}}$ newly assigned to cluster $c$, \workname~compares it with existing personas $p_i \in c$ using an LLM-based semantic judge:
$y=\texttt{LLM}_{\texttt{judge}}(\text{prompt}_{\text{rel}}, p_i, p_{\text{cand}})$,
where $\text{prompt}_{\text{rel}}$ is the prompt. $y$ denotes the semantic relation between the two personas, indicating whether they describe similar, conflicting, or unrelated behavioral patterns.
If the personas are similar, their supporting evidence is merged by updating the evidence timestamps.
If they conflict, both personas are retained within the cluster, and their temporal weights are adjusted as described in Sec.~\ref{sec:persona_updating}. 
Otherwise, the candidate persona is added as a new persona within the cluster.

\vspace{-.5em}

\subsubsection{Temporal Evidence-Aware Persona Updating}
\label{sec:persona_updating}

User personas evolve as routines, environments, and interaction patterns change over time~\cite{roberts2006patterns}. Such variations may introduce new behavioral patterns and lead to conflicting personas. Meanwhile, previously observed personas may remain relevant because similar contexts can recur.
To address this challenge, \workname~is motivated by memory decay theory in cognitive psychology~\cite{wixted2004psychology} and introduces a temporal evidence-aware persona updating mechanism.

\workname~maintains a weight for each persona based on its supporting evidence over time. As described in Sec.~\ref{sec:Inter-Episode Multidimensional Persona Reasoning}, each persona inferred by \workname~is associated with traceable evidence, including supporting episodic events and their timestamps. These timestamps enable the system to estimate the temporal relevance of each persona.
Specifically, \workname~computes the weight of persona $p$ at time $t$ as $w(p,t)=|\mathcal{T}(p)|\exp(-(t-t_{last})/\gamma)$, where $t_{last}$ denotes the timestamp of the most recent supporting evidence for persona $p$, $|\mathcal{T}(p)|$ denotes the number of supporting episodic events, and $\gamma$ controls the temporal decay rate.
Personas without recent evidence progressively decay in weight over time. If a previously observed behavioral pattern reappears, the newly observed evidence immediately increases the weight of the corresponding persona, enabling rapid reactivation. 
In this work, we set $\gamma=30$ days, corresponding to a monthly decay horizon. 
Personas not supported by new evidence for an extended period are removed from the database.

%% file: secs/5_evaluation.tex
\section{Evaluation}
\subsection{Data Collection}
\label{sec:data_collection}
To the best of our knowledge, no existing dataset provides in-the-wild, longitudinal, multimodal sensor streams spanning weeks to months with persona annotations.  
We therefore collect a real-world dataset to evaluate \workname. 


We recruit 20 participants (10 male, 10 female), aged 18 to 63 years (mean $31.4 \pm 12.6$), with diverse occupational backgrounds and daily routines, including undergraduate, master's, and PhD students from various majors, as well as working professionals and retired adults.
This study was approved by the institution’s IRB, and all participants provided informed consent prior to data collection.
Each participant used their own smartphone with our data collection app installed and continued their normal daily routines.
The app continuously recorded seven phone-sensed signals, including audio, IMU, GPS, network information, battery level, screen brightness, and pedometer data.
Participants could pause data collection at any time for privacy concerns.
Otherwise, data collection ran continuously throughout the day except during sleep.
After data collection, each participant completed a questionnaire to self-report their personas. 
To mitigate incomplete or uncertain self-descriptions, we provided a guideline suggesting several reference aspects (e.g., physical routines and interaction patterns) with example persona descriptions from prior work~\cite{ge2024scaling}.
Participants were not limited in the number of personas they could report.


Tab.~\ref{tab:dataset_stats} shows the statistics of the self-collected dataset.
The dataset spans 17 cities across three continents over periods of up to three months, totaling 1,580 hours of sensor recordings.
Personas descriptions average 289 words per participant, with up to 22 personas per participant.
Among the participants, eight exhibited noticeable changes in location and daily routines during the data collection period. Specifically, they first spent 3–4 weeks in a study or work phase in one city, then traveled to their hometown for a winter break of two weeks, during which their locations, physical patterns, social interactions, and conversation topics changed significantly. 
After the break, they returned to the original city and resumed their regular routines.

\begin{table}[t]
\centering
\footnotesize
\caption{Detailed statistics of the self-collected dataset.}
\label{tab:dataset_stats}
\vspace{-1.3em}
\begin{tabularx}{\columnwidth}{p{1.2cm} p{0.6cm} X}
\toprule
\textbf{Category} & \textbf{Count} & \textbf{Details} \\
\midrule

Participants 
& 20 
& 10 male, 10 female. \\

Age 
& N.A. 
& 18--63 years old (mean $31.4 \pm 12.6$). \\

Cities 
& 17 
& 
The cities\footnotemark~span three continents (Europe, North America, and Asia), with inter-city distances ranging from about 27 km to 12,955 km.
\\

Duration 
& N.A. 
& 3 months, totaling 1580 hours of multimodal sensor streams from smartphones. \\

Modalities 
& 7 
& GPS, Audio, IMU, Network (WiFi SSID, Cellular), battery level, step counts, screen brightness.\\

Smartphone
& 13 
& iPhone (8–17 series), Pixel 7, Huawei Mate 40, OnePlus Ace 3 Pro, Xiaomi 12S Ultra, and vivo S17 Pro \\

\bottomrule
\end{tabularx}

\end{table}
\footnotetext{Specific city names in our self-collected dataset are anonymized throughout the paper for double-blind review.}



\subsection{Task Description}
We comprehensively evaluate \workname~across three tasks, including persona extraction quality, persona-aware agent responses, and a user study with human ratings.


\subsubsection{Persona Extraction}
We first evaluate the quality of personas extracted by \workname~against participants' self-reported ground truth. 
Following prior work~\cite{chen2025halumem,wang2024ai,zhong2024memorybank}, we use \textit{Recall}, \textit{Precision}, and \textit{F1} for evaluation.

\noindent\textbf{Recall}.
This metric measures the proportion of self-reported personas that are correctly predicted. 
Following prior work~\cite{chen2025halumem}, we use an LLM-as-judge to determine whether each self-reported persona aligns with at least one inferred persona.

\noindent\textbf{Precision}.
Following prior work~\cite{chen2025halumem,wang2024ai,zhong2024memorybank}, we adopt an LLM-as-judge to assess each inferred persona, evaluating whether it is accurate rather than hallucinated.



\noindent\textbf{F1}.
This is the harmonic mean of recall and precision.


\subsubsection{Persona-Aware Agent Response}
Next, we evaluate the effectiveness of equipping off-the-shelf LLM agents, such as ChatGPT~\cite{openai_chatgpt_2023}, with personas inferred by \workname.
We randomly select 50 user queries from public LLM assistant datasets covering common daily use cases (e.g., scheduling and recommendations)~\cite{jiang2025know,zhao2025llmsrecognize}.
Personas derived from different approaches serve as external memory accessible to the agent during inference.   
We then evaluate the quality of the agent’s responses under different persona sources. Following prior work~\cite{wang2024ai}, we use an LLM-as-judge to assess responses across four dimensions: helpfulness, correctness, completeness, and actionability. 
We conduct pairwise comparisons between responses from \workname~and each baseline, reporting win, tie, and loss rates.
To avoid confounding effects from prior conversations, we disabled the agent’s built-in memory during evaluation.
Following prior work~\cite{wang2024ai}, we evaluate only the agent’s first response to each query.


\subsubsection{Human Evaluation}

Self-reported personas may be incomplete, as participants can overlook some of their own preferences and psychosocial traits. 
We therefore conduct a human evaluation by asking each participant to rate the personas inferred by \workname~(see \S~\ref{sec:user_study}).
Notably, low \textit{Precision} paired with high human ratings would suggest that \workname~ identifies valid implicit personas, highlighting its ability to uncover traits users may not be aware of.






\subsection{System Implementation}

\subsubsection{Testbed Setup}

We implement \workname~on a testbed consisting of a smartphone and a backend server with 8$\times$ NVIDIA RTX A6000 GPUs. 
The smartphone collects seven types of sensor data, including GPS, IMU, audio, step count, battery level, screen brightness, and network connectivity.
Inertial sensors are sampled at 100 Hz, while location data are recorded at 1-second intervals. 
Audio is offloaded only when speech is detected using a voice activity detection mechanism, reducing the average data volume from 12.53 MB/h to 2.42 MB/h.
We set $\alpha$, $\theta$, and $T$ to 0.3, 0.65, and 0.8.


\vspace{-.3em}
\subsubsection{Baselines}
We select six strong baselines for comparison, spanning sensor context understanding~\cite{post2025contextllm,xu2025autolife}, long context compression~\cite{jiang2023llmlingua,jiang2024longllmlingua}, and agent memory~\cite{chhikara2025mem0,liu2026simplemem}.

\noindent\textbf{ContextLLM~\cite{post2025contextllm}}.
This approach utilizes LLM-based sensor-context reasoning to derive abstract insights from multi-sensor streams. We adapt the original prompt to infer personas directly from raw multimodal sensor contexts.

\noindent\textbf{AutoLife~\cite{xu2025autolife}}.
It uses LLMs to interpret sensor contexts such as motion and location to generate daily life journals. Audio data is excluded. We adapt the prompt to infer personas from daily sensor data and aggregate them across multiple days.

\noindent\textbf{LLMLingua~\cite{jiang2023llmlingua}}.
It compresses prompts via a coarse-to-fine pipeline that uses a small language model to score token importance and iteratively prunes low-information tokens.


\noindent\textbf{LongLLMLingua~\cite{jiang2024longllmlingua}}.
It extends LLMLingua to long-context scenarios by preserving salient segments while compressing less relevant tokens to meet LLM context limits.



\noindent\textbf{Mem0~\cite{chhikara2025mem0}}.
This approach extracts daily memory items, which capture salient, reusable facts and preferences from the day’s speech, then synthesizes them across days to produce personas from the aggregated memory items.

\noindent\textbf{SimpleMem~\cite{liu2026simplemem}}.
This baseline extracts daily memory items from each day’s speech content.
It employs a write-time novelty gating step that filters redundant items based on semantic similarity before synthesizing cross-day personas. 

Since Mem0 and SimpleMem are designed for textual dialogues rather than multimodal sensor signals, we transcribe the audio streams into text for their processing.
For the persona-aware agent response task, we also include two additional baselines: no personas~(No Persona) and personas derived from chatbot conversation histories~(ChatBot).

\vspace{-.3em}
\subsubsection{Models}
We deploy \workname~using eight LLMs, including Claude-Opus-4.6~\cite{claude}, Gemini-3.1-pro-preview~\cite{gemini}, GPT-4o, GPT-4.1, GPT-5.1, and GPT-5.2~\cite{chatgpt}, Qwen2.5-7B~\cite{qwen2.5}, and Qwen3-8B~\cite{qwen3}. We use a two-layer 1D CNN to classify IMU signals as still or moving. We use \texttt{all-MiniLM-L6-v2}~\cite{reimers2019sentence} to compute semantic similarity.
We use Silero for voice activity detection~\cite{Silero_VAD}, Resemblyzer~\cite{Resemblyzer} for speaker diarization, and SenseVoice~\cite{SenseVoice} to extract speech content and emotion.
\workname~uses 19 types of sensor cues (see details in \S~Appendix~\ref{appendix:dataset}). We will open-source code upon publication.

\begin{figure}[t]
  \centering

 \begin{subfigure}{1\columnwidth}
    \centering
    \includegraphics[width=1\columnwidth]{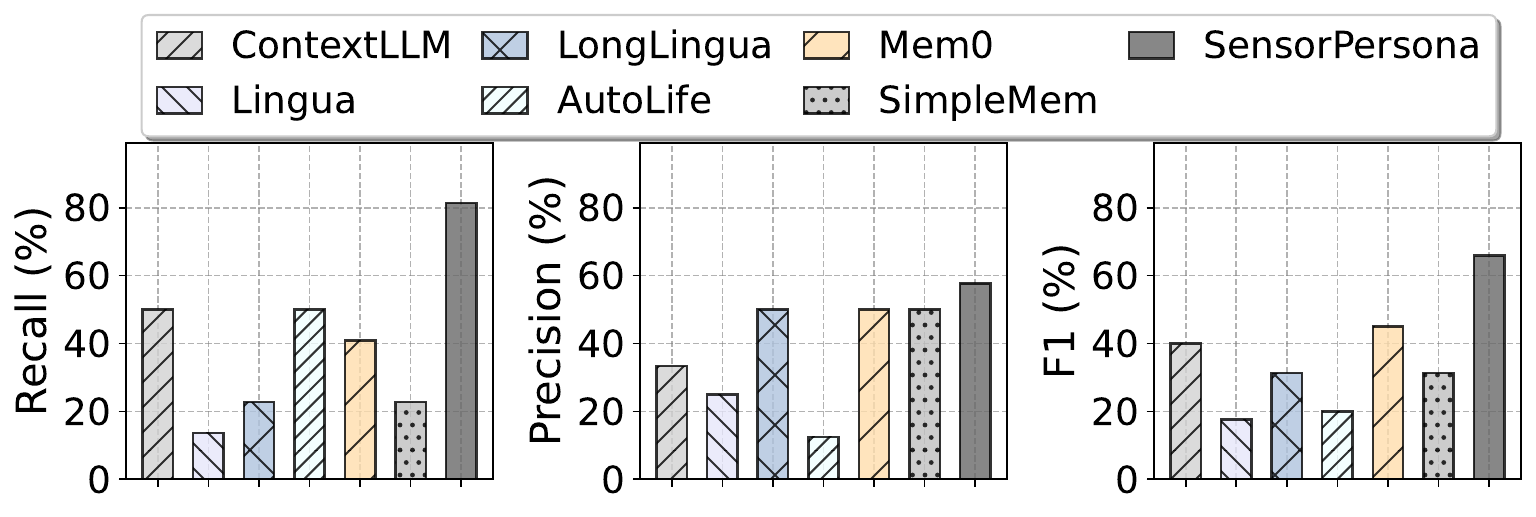}
    \vspace{-2em}
\caption{GPT-4.1.} \label{fig:gpt-4.1}
\end{subfigure}

 \begin{subfigure}{1\columnwidth}
    \centering
    \includegraphics[width=1\columnwidth]{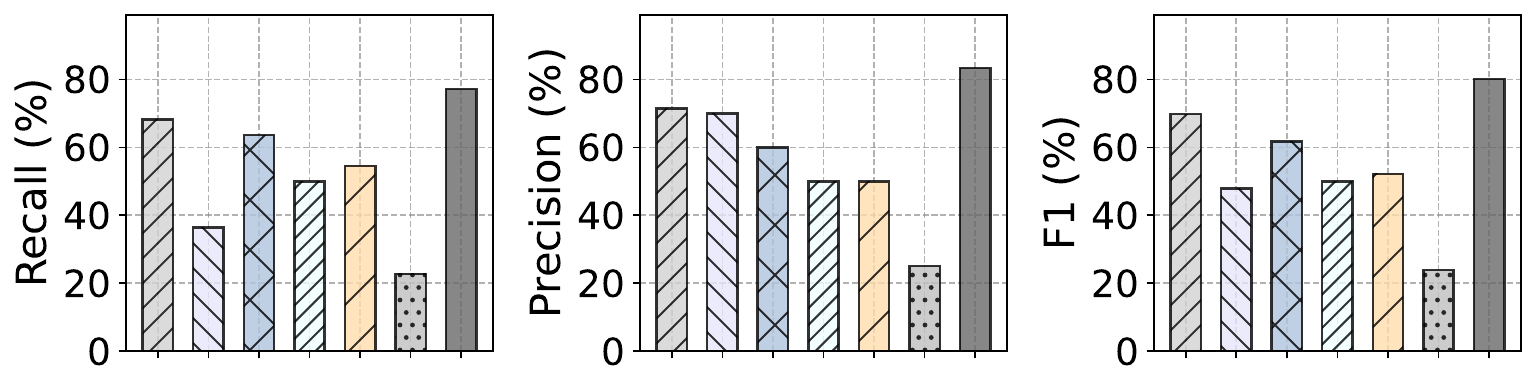}
    \vspace{-2em}
\caption{Claude Opus 4.6.} \label{fig:baselines_claude}
\end{subfigure}

 \begin{subfigure}{1\columnwidth}
    \centering
    \includegraphics[width=1\columnwidth]{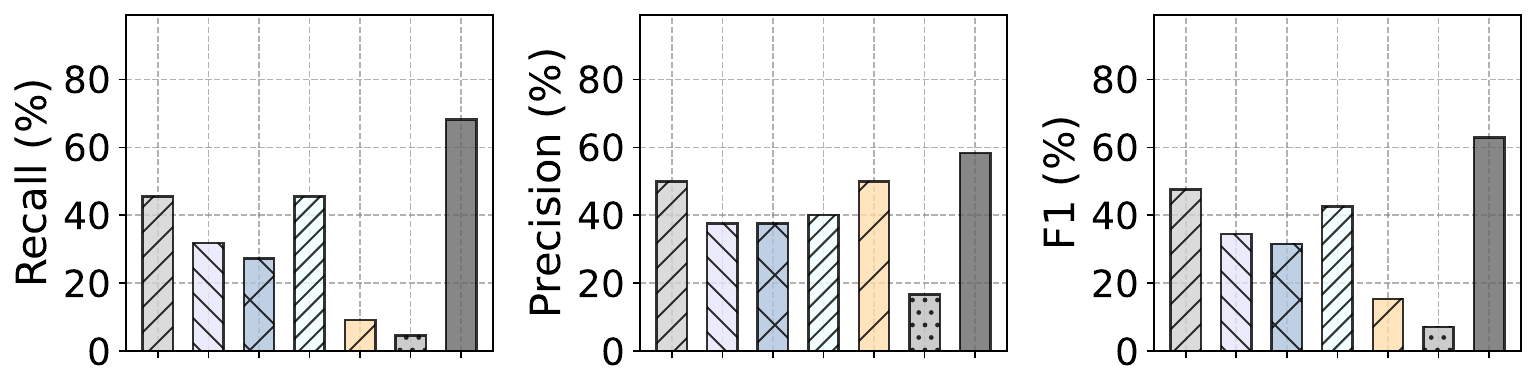}
    \vspace{-2em}
\caption{Qwen3-8B.} \label{fig:baselines_qwen3_8b}
\end{subfigure}
\vspace{-2em}
  \caption{Overall performance of persona extraction.}
\label{fig:overall}
  \vspace{-1.5em}
\end{figure}

\subsection{Overall Performance}

\subsubsection{Quantitative Results}
We first evaluate the persona extraction performance of \workname. Fig.~\ref{fig:overall} demonstrates that \workname~consistently achieves higher \textit{Recall}, \textit{Precision}, and \textit{F1} than the baselines. In particular, \workname~improves \textit{Recall} by up to 31.5\%, suggesting more comprehensive coverage of participants’ self-reported personas. It also improves \textit{Precision} by up to 32.7\%, indicating more accurate persona inference with fewer hallucinations.

\begin{figure}[t]
  \centering
\includegraphics[width=1\linewidth]{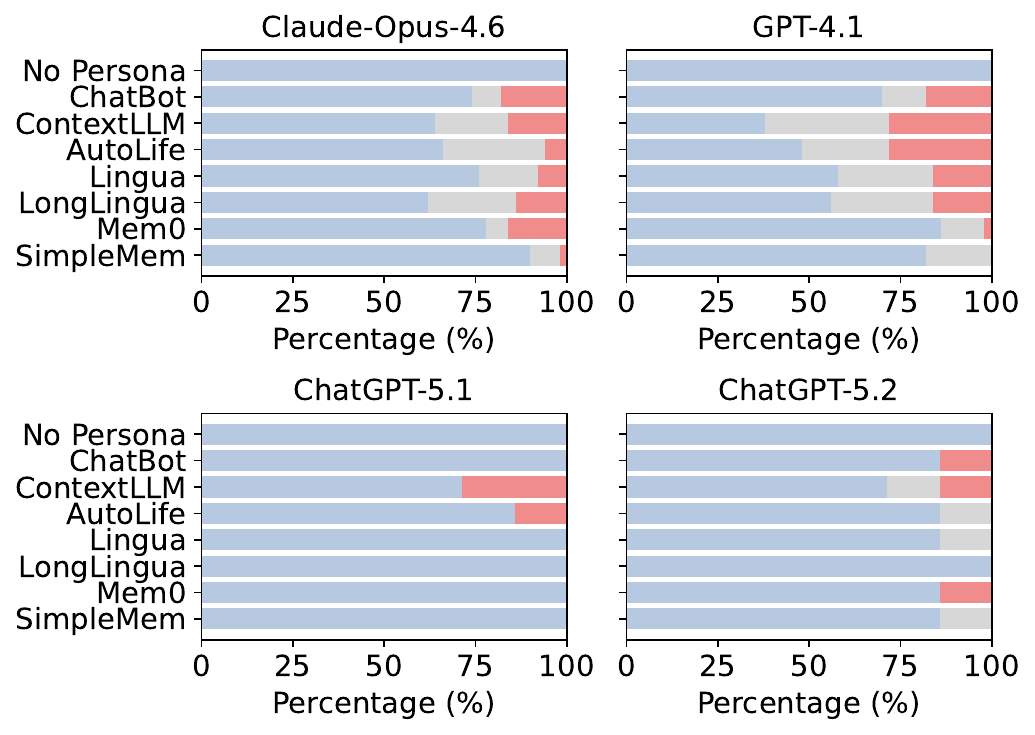}
\vspace{-2em}
  \caption{Overall performance on the persona-aware agent response task. Blue, gray, and red denote win, tie, and loss against the baseline. ``Chatbot'' denotes personas derived from dialogue history.}
  \vspace{-1.3em}
\label{fig:qa_pairwise_stacked}
\end{figure}

Next, we evaluate performance on the persona-aware agent response task.
Fig.~\ref{fig:qa_pairwise_stacked} illustrates the pairwise comparison results between \workname~and the baselines. 
We compare persona sources: no personalized memory (No Persona), memories derived from chatbot dialogue histories (ChatBot), and personas inferred by each sensor-based baseline.
Results show that personas derived from \workname~consistently improve response quality, achieving a 100\% win rate over the first two conditions in most cases and outperforming other sensor-based baselines by up to 85.7\% (with ChatGPT-5.2).
We also analyze performance across different query categories. 
\workname~shows clear advantages over baselines on daily planning and entertainment or social queries, while gains are less pronounced for general knowledge queries (e.g., research or factual questions), which rely less on behavioral patterns captured by sensors.




\begin{figure}[t]
  \centering
\includegraphics[width=1\linewidth]{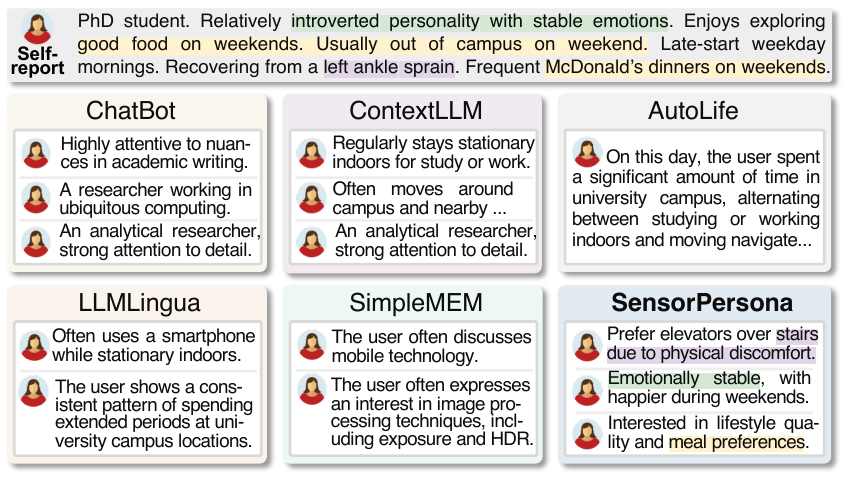}
\vspace{-2em}
\caption{Qualitative results of persona extraction. Only a subset of inferred personas is shown due to space limits. ``Chatbot'' refers to personas derived from the user’s conversation history with ChatGPT.
}
\label{fig:qualitative_result_baselines}
  \vspace{-1.8em}
\end{figure}

\begin{figure}[t]
  \centering
\includegraphics[width=1\linewidth]{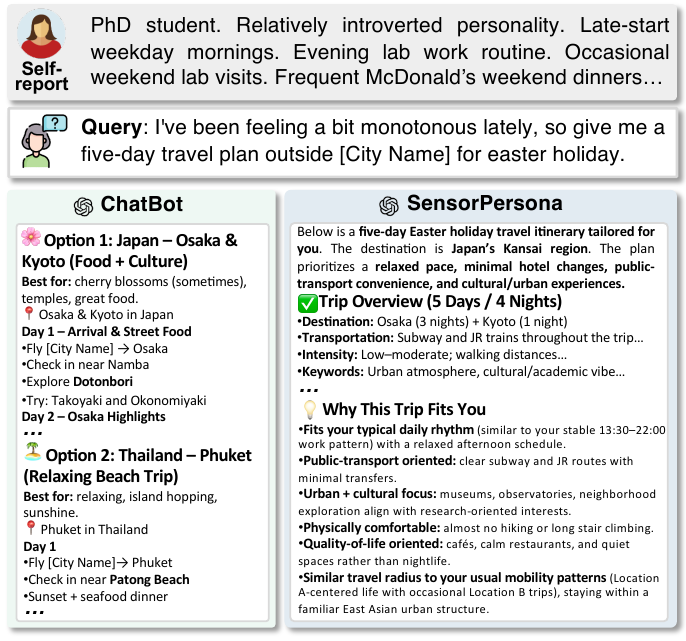}
\vspace{-2em}
\caption{Qualitative results of agent response. The left and right examples show chatbot responses using memories from the ChatGPT-5.2 text box and personas inferred by \workname, respectively.}
  \vspace{-1.2em}
\label{fig:qa_qualitative_results}
\end{figure}



\subsubsection{Qualitative Results}
Fig.~\ref{fig:qualitative_result_baselines} illustrates the qualitative results of the personas derived from \workname~and baselines.
Results show that AutoLife focuses mainly on routine temporal patterns, and many sporadic behaviors it captures are not stable enough to form personas. 
ContextLLM directly uses raw sensor contexts as LLM input, which may span weeks or even months, leading the model to focus on recent moments while diluting persona-relevant cues and producing only a limited set of personas.
Mem0 and SimpleMem rely primarily on self-disclosure from conversation histories, limiting their ability to capture users’ stable physical patterns.
In contrast, \workname~derives personas from longitudinal sensor data, capturing both physical and psychological traits and better aligning with users’ self-reported personas.
Fig.~\ref{fig:qa_qualitative_results} shows examples of agent responses from the chatbot with and without personas derived from \workname.
Results show that integrating personas inferred by \workname~enables the chatbot to generate more personalized responses. In contrast, relying solely on dialogue history leads the chatbot to produce generic plans that provide limited actionable guidance for the user’s query.
%

\vspace{-.8em}
\subsection{Effectiveness of System Module}

\subsubsection{Impact of Context Compression}
\begin{figure}[t]
  \centering
\includegraphics[width=1\linewidth]{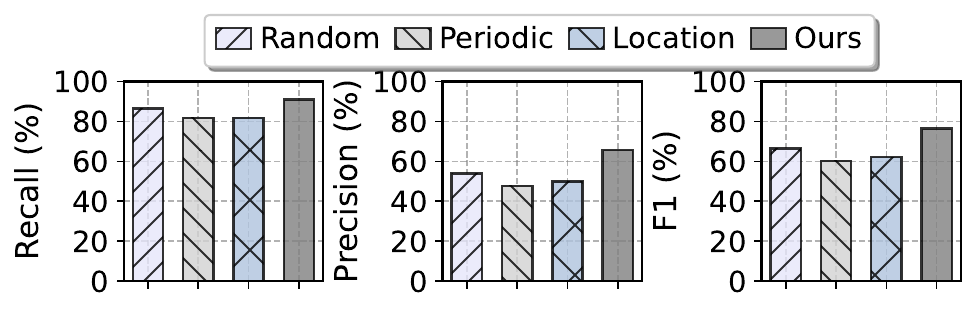}
\vspace{-2.5em}
  \caption{Impact of context compression.}
\label{fig:compression_baselines}
  \vspace{-1.5em}
\end{figure}

We first evaluate the performance of incremental semantic context compression.
Specifically, we compare our approach with several baselines, including random sampling, periodic downsampling, and filtering based solely on attribute similarity from a single sensor context (using location as the attribute in our experiments). We keep the compression rate the same for fairness. Fig.~\ref{fig:compression_baselines} shows that our approach consistently outperforms these baselines, achieving up to 16.1\% higher \textit{F1}, indicating that it preserves more useful persona-related cues.

\subsubsection{Impact of Multi-dimensional Personas}
\begin{figure}[t]
  \centering
\includegraphics[width=1\linewidth]{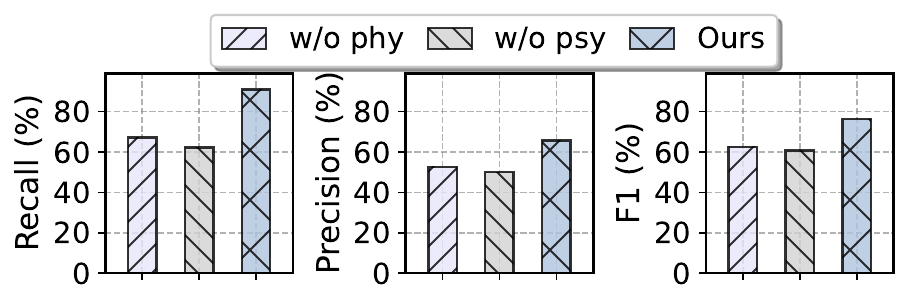}
\vspace{-2em}
  \caption{Impact of multi-dimensional personas. ``w/o phy'' and ``w/o psy'' represent removing physical and psychological personas, respectively.}
  \vspace{-1.em}
\label{fig:ablation_personas}
\end{figure}

We further evaluate the effectiveness of hierarchical multidimensional persona reasoning. 
Fig.~\ref{fig:ablation_personas} illustrates that removing physical and psychological personas reduces the \textit{F1} score by 13.7\%, 15.6\%, respectively, highlighting the importance of these dimensions for comprehensive user understanding.


\begin{figure}[t]
    \centering
    \begin{subfigure}{0.49\columnwidth}  
    \centering \includegraphics[width=1.0\columnwidth]{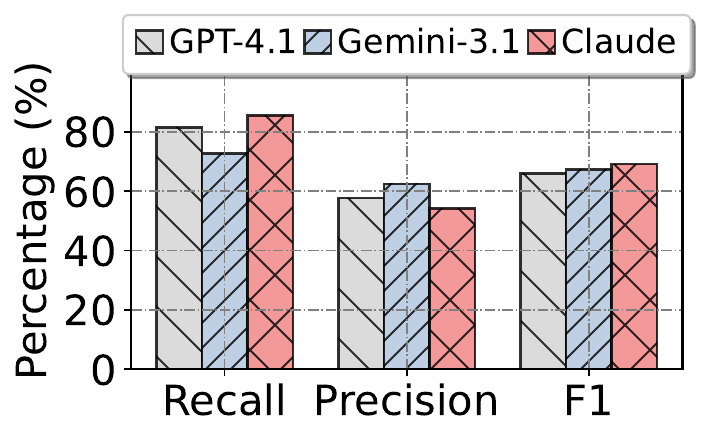}
    \end{subfigure}
    \hfill
     \begin{subfigure}{0.49\columnwidth}
        \centering
        \includegraphics[width=1\columnwidth]{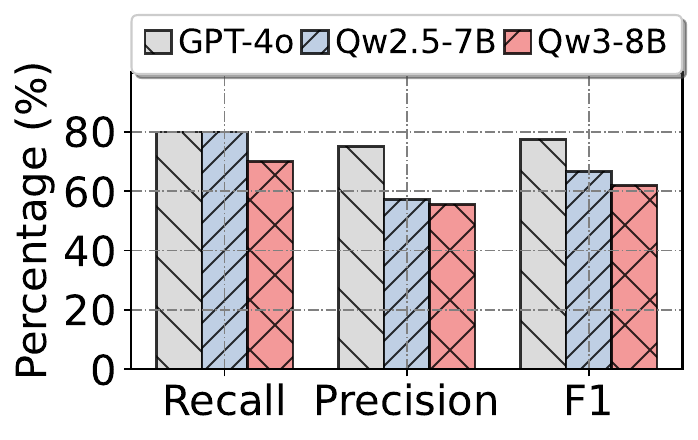}
    \end{subfigure}
   \vspace{-1em} 
\caption{Comparison of different base LLMs used in \workname. Qw refers to Qwen, Claude to Claude Opus 4.6, and Gemini-3.1 to Gemini 3.1 Pro.}
\label{fig:base_llms}
\vspace{-1.5em} 
\end{figure}

\subsubsection{Impact of Persona Maintenance}
\begin{figure}[t]
    \centering
    \begin{subfigure}{0.56\columnwidth}  
    \centering \includegraphics[width=1.0\columnwidth]{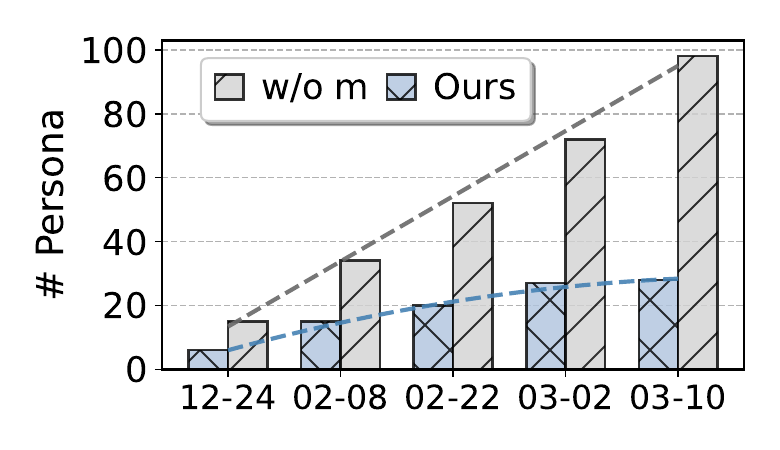}
            \vspace{-2.em} 
    \caption{Number of Personas.} \label{fig:persona_count}
    \end{subfigure}
    \hfill
     \begin{subfigure}{0.43\columnwidth}
        \centering
        \includegraphics[width=1\columnwidth]{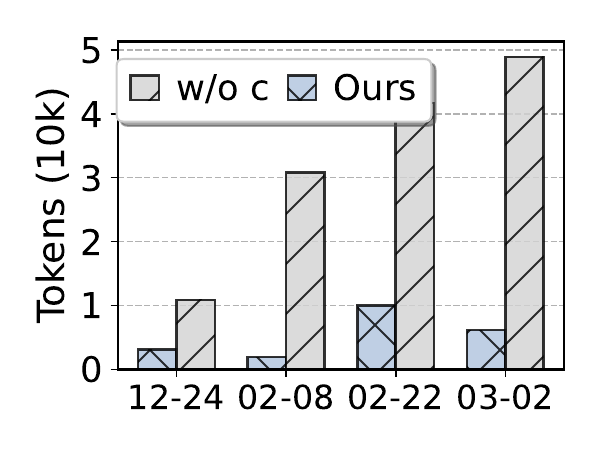}
        \vspace{-2.em} 
        \caption{Token consumption.} \label{fig:token_usage}
    \end{subfigure}
   \vspace{-2em} 
    \caption{Impact of persona maintenance.}
\label{fig:persona_maintenance}
\vspace{-1.5em} 
\end{figure}

\begin{figure}
\begin{minipage}[b]{0.47\columnwidth} 
     \centering
\includegraphics[width=1.05\linewidth]{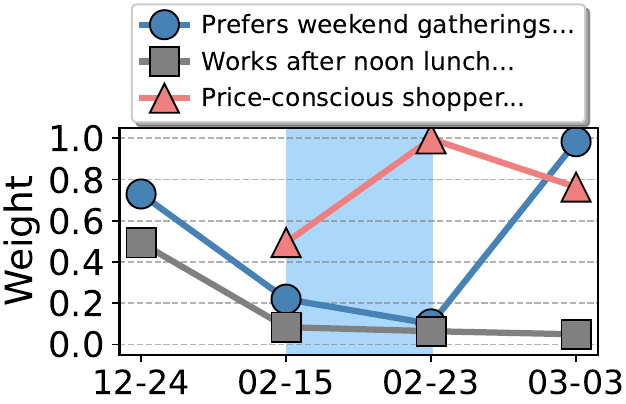}
\vspace{-1.5em}
  \caption{Changes in persona weight over time.}
\label{fig:weight_summary_line}
\end{minipage}
\hfill
  \begin{minipage}[b]{0.5\columnwidth} 
     \centering
\includegraphics[width=1\linewidth]{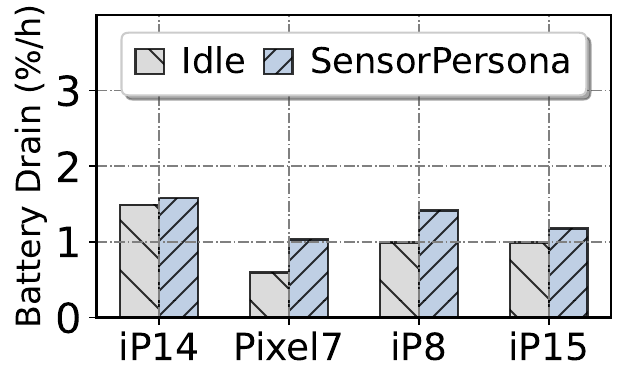}
\vspace{-1.5em}
  \caption{Energy consumption. IP denotes iPhone.}
\label{fig:energy}
\end{minipage}
\vspace{-1.em}
\end{figure}

We then evaluate the effectiveness of our hierarchical persona maintenance mechanism. Fig.~\ref{fig:persona_maintenance} and Fig.~\ref{fig:weight_summary_line} show how the personas extracted by \workname~evolve over a continuous three-month period, including their number, weights, and token consumption.

Specifically, Fig.~\ref{fig:persona_count} shows the number of personas over time. We compare our approach with a baseline without the persona maintenance mechanism (\textit{w/o m}).
Results show that this baseline continuously accumulates personas, whereas \workname~initially increases and then stabilizes, significantly reducing redundant personas.
Moreover, Fig.~\ref{fig:weight_summary_line} shows the evolution of persona weights inferred by \workname~over time, highlighting three representative personas. A location change during Feb.~15–Feb.~23 (winter break) leads to the emergence of several specific personas, reflected by the red curve. The blue persona reappears in March and is quickly reactivated, whereas the gray persona rarely recurs and its weight gradually decays.

Finally, we evaluate the effectiveness of incremental persona clustering. We compare \workname~with a baseline that performs similarity checks without streaming clustering, denoted as \textit{w/o c}. 
Fig.~\ref{fig:token_usage} shows that our approach reduces token consumption by up to 7.9$\times$ compared to this baseline.


\begin{figure}[t]
  \centering
\includegraphics[width=1\linewidth]{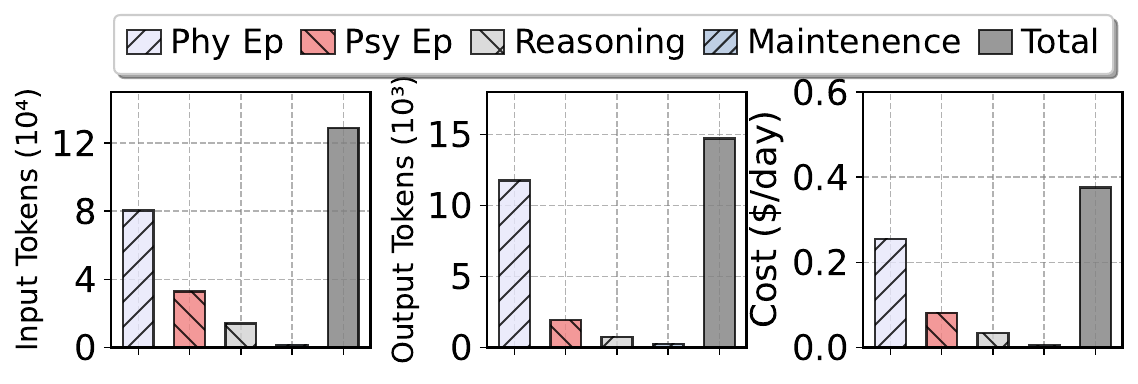}
\vspace{-2em}
  \caption{Token consumption and cost. \textit{Phy Ep} and \textit{Psy Ep} are physical and psychosocial episode construction, respectively. Cost is computed using GPT-4.1 pricing.
  }
  \vspace{-1.em}
\label{fig:token_consumption}
\end{figure}

\subsubsection{Impact of Different Base Model}
Next, we evaluate \workname’s persona extraction performance across different base LLMs. Fig.~\ref{fig:base_llms} shows that stronger proprietary models such as Claude Opus 4.6 achieve substantially better performance than open-source models such as Qwen3-8B, improving \textit{Recall} by up to 17.3\%. We attribute this gap to the complexity of persona inference, which requires long-horizon reasoning over multimodal sensor streams and is more demanding for smaller models.

\subsubsection{Impact of Hyperparameters}
\label{sec:Impact of Hyper-parameters}
\begin{figure}
\begin{minipage}[b]{0.51\columnwidth} 
     \centering
\includegraphics[width=1\linewidth]{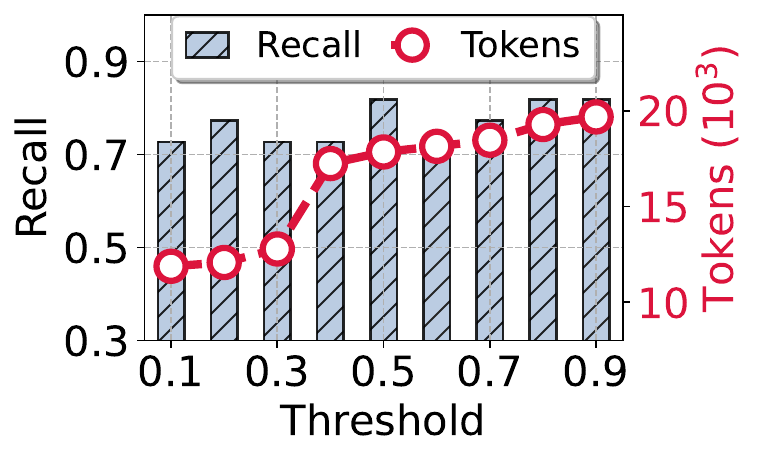}
\vspace{-2em}
  \caption{Impact of $\alpha$.}
\label{fig:tokens_recall_gpt}
\end{minipage}
\hfill
  \begin{minipage}[b]{0.47\columnwidth} 
     \centering
\includegraphics[width=1\linewidth]{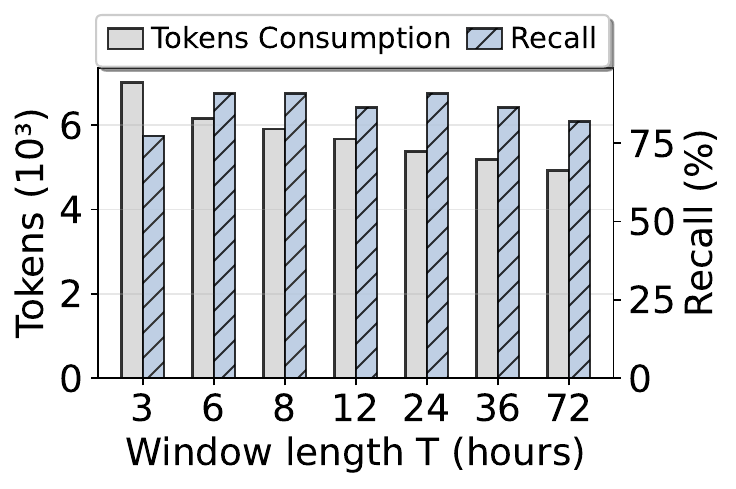}
\vspace{-2em}
  \caption{Impact of $T$.}
\label{fig:time_window_tokens_recall}
\end{minipage}
\vspace{-1.5em}
\end{figure}

We further study the impact of hyperparameters in \workname. 
First, we evaluate the effect of the threshold $\alpha$ in incremental semantic context compression. Fig.~\ref{fig:tokens_recall_gpt} demonstrates that reducing $\alpha$ from 0.4 to 0.3 reduces input tokens by about 25.9\% while causing only a marginal change in \textit{Recall}.
We therefore set $\alpha$ to 0.3.
We also evaluate the impact of the time interval $T$ on persona extraction. 
Fig.~\ref{fig:time_window_tokens_recall} illustrates that token consumption increases with larger $T$. \workname~achieves the highest \textit{Recall} at $T=8$ while maintaining relatively low token consumption. 
Across all settings, \workname~consistently achieves higher \textit{Recall} than the baselines.


\subsection{System Overhead}
We comprehensively evaluate the system overhead of \workname, including energy consumption, token usage, and LLM costs.
Fig.~\ref{fig:token_consumption} shows the token consumption and corresponding cost for one day of system usage. Episode construction accounts for the majority of token usage, while persona inference and maintenance incur relatively small overhead. Overall, processing one day of data consumes about 143K input tokens, corresponding to less than \$1 per day under GPT-4.1 pricing.
Fig.~\ref{fig:energy} shows the energy consumption on the smartphone.
Results show that \workname~increases smartphone battery consumption by an average of 0.3\% per hour, without significantly affecting normal daily usage.





\vspace{-.1em} 
\subsection{User Study}
\label{sec:user_study}
We conducted a user study with 20 participants. The study comprised two parts. In the first part, participants reviewed the personas inferred by \workname~and rated them on a 
5-point Likert scale across five dimensions. \textit{Accuracy} captures whether a persona correctly describes the participant. \textit{Stability} reflects whether it represents long-term, recurring characteristics rather than temporary behaviors. \textit{Coverage} assesses whether it includes important aspects of the participant’s life. \textit{Specificity} measures whether it feels uniquely personal rather than generic. \textit{Clarity} evaluates whether it is clearly expressed and easy to understand.

In the second part, participants tested an agent (ChatGPT-5.2) using personas derived from \workname~and baselines by issuing arbitrary queries and rating the responses. They rated each persona-aware response on a 5-point Likert scale across five dimensions. \textit{Personalization} captures whether the response feels tailored to the participant. \textit{Helpfulness} reflects whether it supports accomplishing the goal. \textit{Relevance} assesses whether it matches the participant’s situation and request. \textit{Actionability} evaluates whether it provides practical, usable suggestions. \textit{Satisfaction} measures the participant’s overall satisfaction with the response.

\begin{figure}[t]
    \centering
    \begin{subfigure}{0.47\columnwidth}  
    \centering \includegraphics[width=1.0\columnwidth]{figs/user_study_1.pdf}
            \vspace{-1.5em} 
    \caption{Persona extraction.} \label{fig:user_study_1}
    \end{subfigure}
    \hfill
     \begin{subfigure}{0.51\columnwidth}
        \centering
        \includegraphics[width=1\columnwidth]{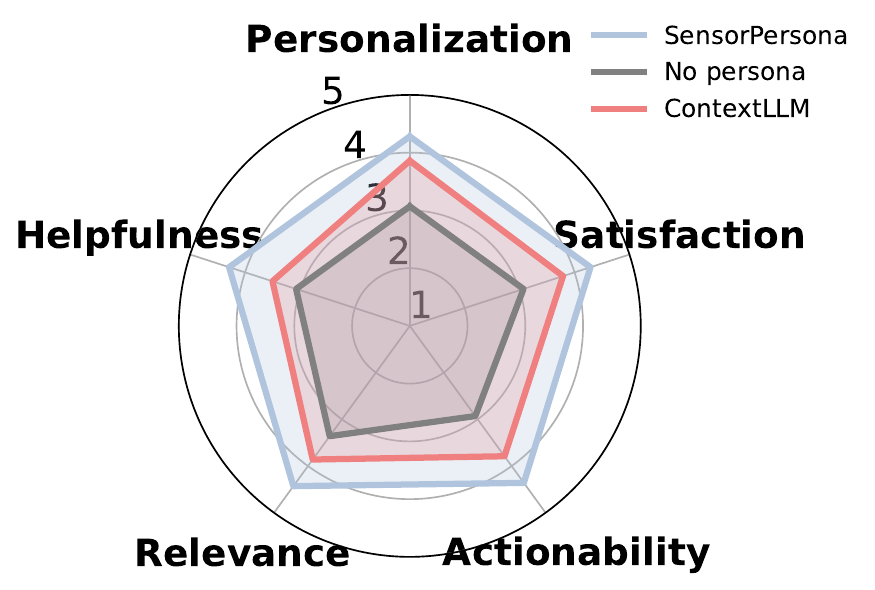}
        \vspace{-1.5em} 
        \caption{Agent response.} \label{fig:user_study_2}
    \end{subfigure}
   \vspace{-1.2em} 
    \caption{User study results of \workname.}
\label{fig:user_study}
\vspace{-1.em} 
\end{figure}

Fig.~\ref{fig:user_study} demonstrates that \workname~consistently outperforms the baselines across different dimensions of persona extraction on both tasks. 
We also analyze ratings from participants who used \workname~for more than 100 hours, totaling 8 participants.
Fig.~\ref{fig:user_study_100_1} and Fig.~\ref{fig:user_study_100_2} further show that when users collect more than 100 hours of sensing data, the average rating of \workname~increases from 4.35 to 4.53, while the baseline decreases from 3.70 to 3.45. 
This suggests that longer sensing durations improve persona understanding and lead to higher ratings for \workname, whereas the baseline does not show the same trend.



\begin{figure}[t]
  \centering
\includegraphics[width=1\linewidth]{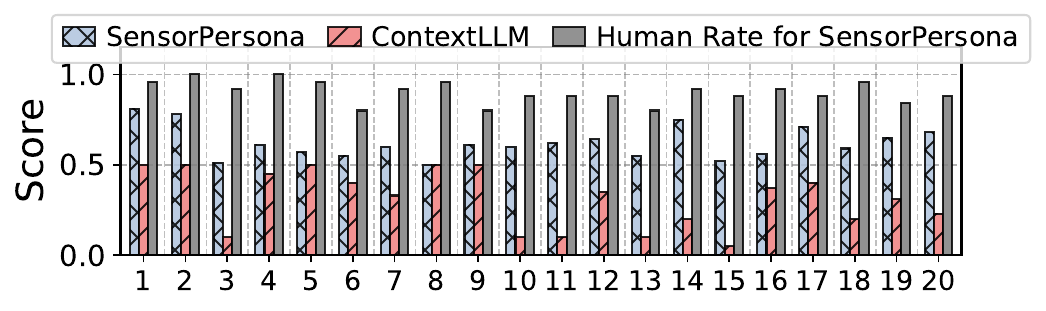}
\vspace{-2.em}
  \caption{User study. Blue and red denote \textit{Precision} scores, while grey is human ratings. Human ratings exceed \textit{Precision}, suggesting that \workname~infers insightful personas users may not explicitly recognize.}
\label{fig:user_study_bar}
  \vspace{-1.em}
\end{figure}

Fig.~\ref{fig:user_study_bar} shows a comparison between \workname~and the baseline in both objective metrics and human ratings.
Results illustrate that some participants with relatively low objective precision still gave \workname~high ratings.
This suggests that the measured \textit{Precision} may partly reflect incomplete self-reported personas rather than system limitations.
Participants also reported that \workname's inferred personas align well with their self-perception, and some noted that \workname~revealed reasonable traits they had not previously recognized, highlighting the value of uncovering latent user personas.
Lower ratings were mainly attributed to recognition errors (e.g., speaker or language misidentification), temporarily disabled sensing modalities (e.g., audio), and insufficient data collection duration, which may hinder capturing stable long-term personas.


%% file: secs/6_discussion.tex
\vspace{-.1em}
\section{Discussion}
\noindent\textbf{Scalability to Personal Agents}.
\workname~can be integrated into personal agents such as OpenClaw~\cite{OpenClaw}. Personas inferred from \workname~complement agent memories derived from chat histories, enabling LLM assistants to better understand users and improve response and task quality.


\noindent\textbf{Privacy Concerns}.
\workname~avoids collecting visual data and stores inferred personas locally on the user’s device, reducing exposure of personal data. Data sent to cloud model APIs is anonymized. \workname~also remains effective with small LLMs like Qwen3-8B, reducing reliance on cloud.



\noindent\textbf{Scalability to Other Sensing Techniques}.
\workname~\\can incorporate more advanced sensing techniques, additional mobile devices, and modalities~\cite{he2025embodiedsense,yang2022novel,yang2023brainz} to enrich the sensory context and improve persona inference.